\documentclass[11pt,a4paper]{article}
\usepackage[hyperref]{acl2021}
\usepackage{times}
\usepackage{latexsym}

\usepackage{microtype}
\usepackage{times}
\usepackage{latexsym}
\usepackage{amsmath}
\usepackage{amssymb}
\usepackage{amsthm}
\usepackage[T1]{fontenc}
\usepackage[utf8]{inputenc}
\usepackage{microtype}
\usepackage{booktabs}
\usepackage{threeparttable}
\usepackage{amsmath}
\usepackage{booktabs}
\usepackage{threeparttable}
\usepackage{multirow}
\usepackage{arydshln}
\usepackage{subcaption}
\usepackage{array}
\usepackage{setspace}
\usepackage{algorithm}
\usepackage{algpseudocode}
\usepackage{bm}
\usepackage[inline]{enumitem}
\usepackage{amssymb}
\usepackage{soul}
\usepackage{tikz}
\usetikzlibrary{calc}
\usepackage{xparse}
\usepackage{bbm}
\usepackage[inline]{enumitem}

\definecolor{orange2}{rgb}{0.77254902, 0.352941176, 0.066666667}
\definecolor{yellow2}{rgb}{0.749019608, 0.564705882, 0.0}
\definecolor{red}{rgb}{1.0,0.0,0.0}
\definecolor{orange}{rgb}{1.0,0.65,0.0}
\definecolor{green}{rgb}{0.0,0.75,0.0}
\definecolor{blue}{rgb}{0.0,0.0,1.0}
\definecolor{purple}{rgb}{0.5,0.0,0.5}

\newlength\LineWidth
\definecolor{HLcolor}{RGB}{124,18,18}
\sethlcolor{HLcolor}

\aclfinalcopy
\def\aclpaperid{562}

\makeatletter

\newcommand{\defhighlighter}[3][]{%
  \tikzset{every highlighter/.style={draw=#2, fill opacity=#3, #1}}%
}

\defhighlighter[fill=olive!15]{HLcolor}{.25}

\newcommand{\highlight@DoHighlight}{
  \fill [outer sep = -15pt, inner sep = 0pt, every highlighter, this highlighter,draw=none]
        ($(begin highlight)+(0,8pt)$) rectangle ($(end highlight)+(0,-2pt)$) ;
  \draw[HLcolor,line width=\LineWidth]  ($(begin highlight)+(0,-2pt)$) -- ($(end highlight)+(0,-2pt)$) ;
  \draw[HLcolor,line width=\LineWidth]  ($(begin highlight)+(0,8pt)$) -- ($(end highlight)+(0,8pt)$) ;
}

\newcommand{\highlight@BeginHighlight}{
  \coordinate (begin highlight) at (0,0) ;
}

\newcommand{\highlight@EndHighlight}{
  \coordinate (end highlight) at (0,0) ;
}

\newdimen\highlight@previous
\newdimen\highlight@current

\DeclareRobustCommand*\highlight[1][]{%
  \tikzset{this highlighter/.style={#1}}%
  \SOUL@setup
  \def\SOUL@preamble{%
    \begin{tikzpicture}[overlay, remember picture]
      \highlight@BeginHighlight
      \draw[HLcolor,line width=\LineWidth]  ($(begin highlight)+(0,-2pt)+(0,-0.5\pgflinewidth)$) -- ($(begin highlight)+(0,8pt)+(0,0.5\pgflinewidth)$) ;
      \highlight@EndHighlight
    \end{tikzpicture}%
  }%
  \def\SOUL@postamble{%
    \begin{tikzpicture}[overlay, remember picture]
      \highlight@EndHighlight
      \highlight@DoHighlight
      \draw[HLcolor,line width=\LineWidth]  ($(end highlight)+(0,-2pt)+(0,-0.5\pgflinewidth)$) -- ($(end highlight)+(0,8pt)+(0,0.5\pgflinewidth)$) ;
    \end{tikzpicture}%
  }%
  \def\SOUL@everyhyphen{%
    \discretionary{%
      \SOUL@setkern\SOUL@hyphkern
      \SOUL@sethyphenchar
      \tikz[overlay, remember picture] \highlight@EndHighlight ;%
    }{%
    }{%
      \SOUL@setkern\SOUL@charkern
    }%
  }%
  \def\SOUL@everyexhyphen##1{%
    \SOUL@setkern\SOUL@hyphkern
    \hbox{##1}%
    \discretionary{%
      \tikz[overlay, remember picture] \highlight@EndHighlight ;%
    }{%
    }{%
      \SOUL@setkern\SOUL@charkern
    }%
  }%
  \def\SOUL@everysyllable{%
    \begin{tikzpicture}[overlay, remember picture]
      \path let \p0 = (begin highlight), \p1 = (0,0) in \pgfextra
        \global\highlight@previous=\y0
        \global\highlight@current =\y1
      \endpgfextra (0,0) ;
      \ifdim\highlight@current < \highlight@previous
        \highlight@DoHighlight
        \highlight@BeginHighlight
      \fi
    \end{tikzpicture}%
    \the\SOUL@syllable
    \tikz[overlay, remember picture] \highlight@EndHighlight ;%
  }%
  \SOUL@
}

\def\adl@drawiv#1#2#3{%
        \hskip.5\tabcolsep
        \xleaders#3{#2.5\@tempdimb #1{1}#2.5\@tempdimb}%
                #2\z@ plus1fil minus1fil\relax
        \hskip.5\tabcolsep}
\newcommand{\cdashlinelr}[1]{%
  \noalign{\vskip\aboverulesep
           \global\let\@dashdrawstore\adl@draw
           \global\let\adl@draw\adl@drawiv}
  \cdashline{#1}
  \noalign{\global\let\adl@draw\@dashdrawstore
           \vskip\belowrulesep}}

\makeatother

\DeclareDocumentCommand\MyDBox{O{HLcolor!15}O{HLcolor}m}{%
  \colorlet{HLcolor}{#1!75}
  \highlight[#1!75]{#3}%
}

\title{Rationalization through Concepts}

\author{Diego Antognini \and Boi Faltings\\
  École Polytechnique Fédérale de Lausanne, Switzerland\\
  \texttt{firstname.lastname@epfl.ch}
}

\date{}

\begin{document}
\maketitle
\begin{abstract}

Automated predictions require explanations to be interpretable by humans.
One type of explanation is a rationale, i.e., a selection of input features such as relevant text snippets from which the model computes the outcome. However, a single overall selection does not provide a complete explanation, e.g., weighing several aspects for decisions.
To this end, we present a novel self-interpretable model called ConRAT. Inspired by how human explanations for high-level decisions are often based on key concepts, ConRAT extracts a set of text snippets as concepts and infers which ones are described in the document. Then, it explains the outcome with a linear aggregation of concepts. Two regularizers drive ConRAT to build interpretable concepts. In addition, we propose two techniques to boost the rationale and predictive performance further. Experiments on both single- and multi-aspect sentiment classification tasks show that ConRAT is the first to generate concepts that align with human rationalization while using only the overall label. Further, it outperforms state-of-the-art methods trained on each aspect label independently.

\end{abstract}

\section{Introduction}

Neural models have become the standard for many tasks, owing to their large performance gains. However, their adoption in decision-critical fields is more limited because of their lack of interpretability, particularly with textual data.

One of the simplest means of explaining predictions of complex models is by selecting relevant input features. Attention mechanisms \cite{iclr2015} model the selection using a conditional importance distribution over the inputs, but the resulting explanations are noisy \cite{jain2019attention,pruthi-etal-2020-learning}. Multi-head attention \cite{vaswani2017attention} extends attention mechanisms to attend information from different perspectives jointly. However, no explicit mechanisms guarantee a logical connection between different views \cite{voita-etal-2019-analyzing,kovaleva-etal-2019-revealing}.
Another line of research includes rationale generation methods \cite{lei-etal-2016-rationalizing,chang2020invariant,antognini2019multi}. If the selected~text input features are short and concise -- called a rationale -- and suffice on their own to yield the prediction, it can potentially be understood and verified against domain knowledge \cite{chang2019game}. 

\begin{figure}[t]
\centering
\includegraphics[width=\linewidth]{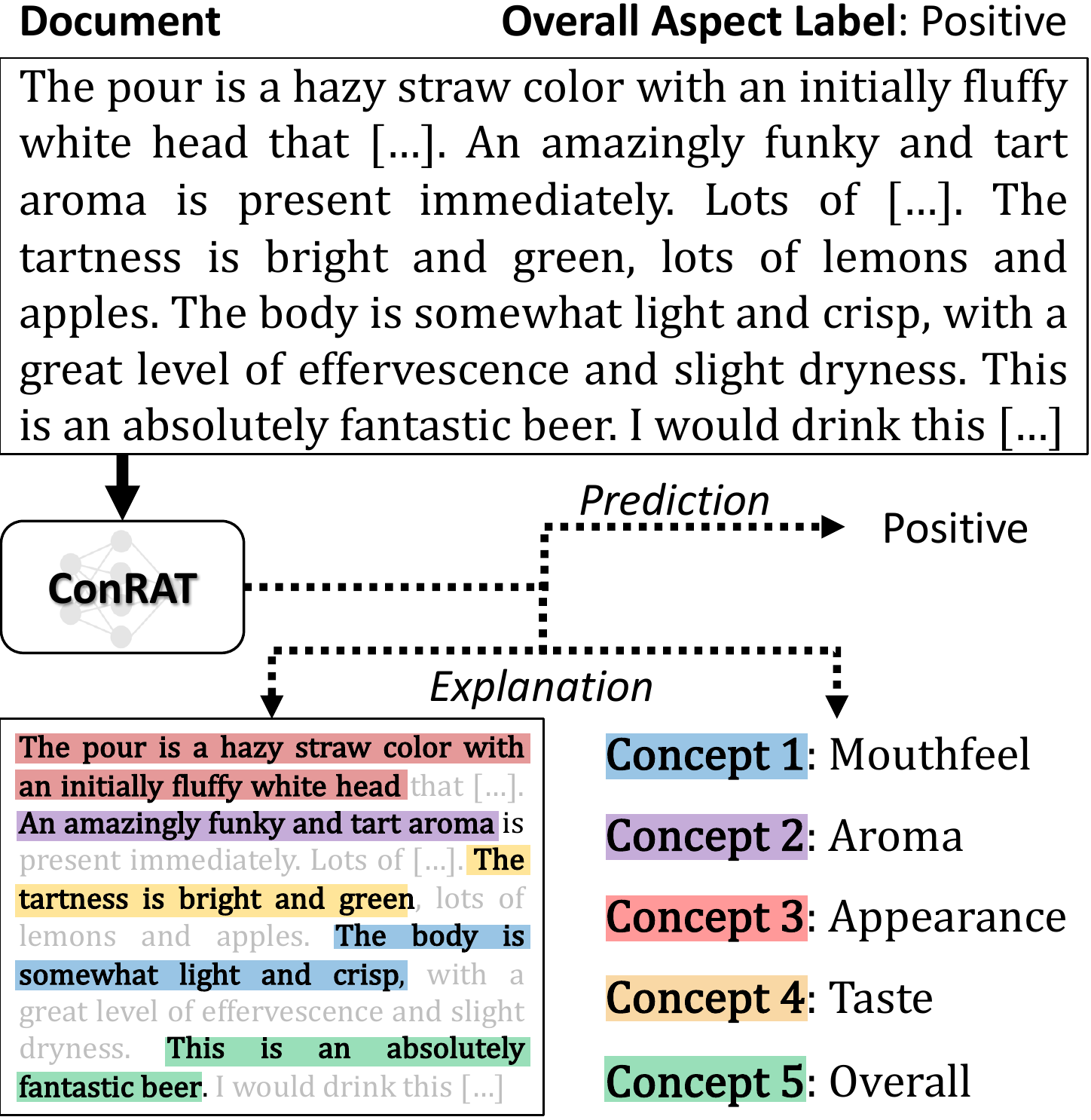}
\caption{\label{front}An illustration of ConRAT. Given a beer~review, ConRAT identifies five excerpts that relate to~particular concepts of beers (i.e., the explanation), depicted in color, from which it computes the outcome.}
\end{figure}

The key motivation for this work arises from the limitations of rationales. Rationalization models strive for one overall selection to explain the outcome by maximizing the mutual information between the rationale and the label. However, useful rationales can be multi-faceted, where each facet relates to a particular ``concept'' (see Figure~\ref{front}). For example, users typically justify their opinions of a product by weighing explanations: one for each aspect they care about \cite{musat2015personalizing}.

Inspired by how human reasoning comprises concept-based thinking \cite{ARMSTRONG1983263,tenenbaum1999bayesian}, we aim to discover, in an unsupervised manner, a set of concepts to explain the outcome with a weighted average, similar to multi-head attention. In this work, we relate concepts to semantically meaningful and consistent excerpts across multiple texts. Unlike topic modeling, where documents are described by a set of latent topics comprising word distributions, our latent concepts relate to text snippets that are relevant for the prediction.

Another motivation for this study is to generate interpretable concepts. The explanation~of~an outcome should rely on concepts that~satisfy the desiderata introduced in \citet{Alvarez-MelisJ18}. They should \begin{enumerate*}
  \item preserve relevant information,
  \item not overlap with each other and be diverse, and
  \item be human-understandable
\end{enumerate*}. Figure~\ref{front} shows an example of concepts in the beer domain.

In this work, we present a novel self-explaining neural model: the concept-based rationalizer (ConRAT) (see Figure~\ref{front} and \ref{architecture}). Our new rationalization scheme first identifies a set of concepts in a document and then decides which ones are currently described (binary selection). ConRAT explains the prediction with a linear aggregation of concepts. The model is trained end-to-end, and the concepts are learned in an unsupervised manner. In addition,~we~design two regularizers that guide ConRAT to induce interpretable concepts and propose two optional techniques, knowledge distillation and concept pruning, in order to boost the performance further.

We evaluate ConRAT on both single- and multi-aspect sentiment classification with up to five target labels. Upon training ConRAT only on the overall aspect, the results show that ConRAT generates concepts that are relevant, diverse, and non-overlapping, and they also recover human-defined concepts. Furthermore, our model significantly outperforms strong supervised baseline models in terms of predictive and explanation performance.

\section{Related Work}

Developing interpretable models is of considerable interest to the broader research community. Researchers have investigated many approaches to improve the interpretability of neural networks. 

\subsection{Interpretability.}
The first line of research aims at providing post-hoc explanations of an already trained model. For example, gradient and perturbation-based methods attribute the decision to important input features \cite{10.1145/2939672.2939778,pmlr-v70-sundararajan17a,NIPS2017_8a20a862,pmlr-v70-shrikumar17a}.
Other studies identified the causal relationships between input-output pairs \cite{alvarez-melis-jaakkola-2017-causal,goyal2019explaining}. In contrast, our model is inherently interpretable as it directly produces the prediction with an explanation.

Another line of research has developed interpretable models. \citet{quint2018interpretable} extended a variational auto-encoder with a differentiable decision tree. \citet{alaniz2019explainable} proposed an explainable observer-classifier framework whose predictions can be exposed as a binary tree. However, these methods have been designed for images only, while our work focuses on text input.

The works most relevant to ours relate to interpretable models from the rationalization field~\cite{lei-etal-2016-rationalizing,bastings-etal-2019-interpretable,yu-etal-2019-rethinking,chang2020invariant,jain-etal-2020-learning,paranjape-etal-2020-information}. These methods justify their predictions by selecting rationales (i.e., relevant tokens in the input text). However, they are limited to explain~only~the prediction with mostly one text span and rely on the assumption that the data have low internal correlations \cite{antognini2019multi}. \citet{chang2019game} extended previous methods to extract an additional rationale in order to counter the prediction. In our work, ConRAT produces multi-faceted rationales and explains the prediction through a linear aggregation of the extracted concepts. However, if we set the number of concepts to one, ConRAT reduces to a special case of a rationale model.

\subsection{Explanations through Concepts.}

Researchers have proposed multiple approaches for concept-based explanations. \citet{conf/icml/KimWGCWVS18} designed a post-hoc technique to learn concept activation vectors by relying on human annotations that characterize concepts of interest. Similarly, \citet{bau2017network,10.1007/978-3-030-01237-3_8} generated visual explanations for a classifier. Our concepts are learned in an unsupervised manner and not defined a~priori. 

Few studies have learned concepts on images in an unsupervised fashion. \citet{li2018deep} explained predictions based on the similarity of the input to ``prototypes'' learned during training. \citet{Alvarez-MelisJ18} used an auto-encoder to extract relevant concepts and explain the prediction. \citet{ghorbani2019towards} designed an unsupervised concept discovery method to explain trained models. \citet{koh2020concept} employed the discovered concepts to predict the target label. Our work's key difference is that we focus on text data, while all these methods treat only image inputs.

To the best of our knowledge, \citet{bouchacourt2019educe} is the only study that has proposed a self-interpretable concept-based model for text data using reinforcement learning. It computes the predictions and provides an explanation in terms of the presence or~absence of concepts in the input (i.e., text excerpts of variable lengths). However, their method achieves poor overall performance. In addition, it is unclear whether the discovered concepts are interpretable. Conversely, ConRAT is differentiable, clearly outperforms strong models in terms of predictive and explanation performance, and it infers relevant, diverse, non-overlapping, and human-understandable concepts. 

\subsection{Topic Modeling.}
Topic models, such as latent Dirichlet allocation \cite{blei2003latent}, describe documents with a mixture of latent topics. Each topic represents a word distribution. Some studies combined topic models with recurrent neural~models \cite{dieng2016topicrnn,zaheer2017latent}. However, the goal of these generative models and the topics remains different than this work's.~We aim to build a self-interpretable model that predicts and explains the outcome with latent concepts.

\section{Concept-based Rationalizer (ConRAT)}

\begin{figure}[t]
\centering
\includegraphics[width=\linewidth]{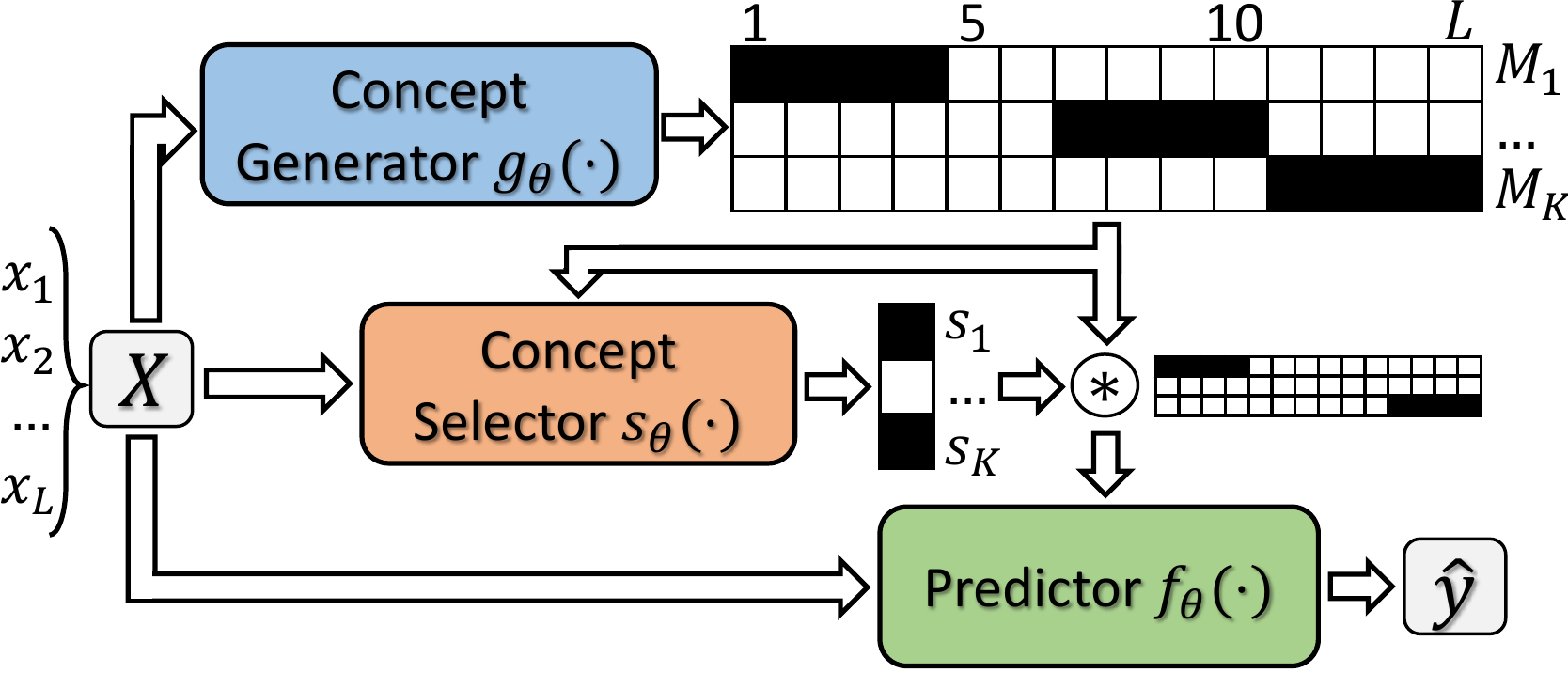}
\caption{\label{architecture}The proposed self-explaining model ConRAT. The model predicts and explains~$\hat{y}$. Given a document~$X$, the concept generator produces one binary mask per concept. The concept selector decides which concepts are present in the input. The predictor~aggregates each selected concept's prediction to~compute~$\hat{y}$.}
\end{figure}

Figure~\ref{architecture} depicts the architecture of our proposed self-explaining model: the Concept-based Rationalizer (ConRAT). Let $X$ be a random variable representing a document composed of $T$ words $(x_1, x_2, \dots, x_T)$, $y$ the ground-truth label, and $K$ the desired numbers of concepts.\footnote{Our method is easily adapted for regression problems.}
 Given a document $X$ and a label $y$, our goal is to explain the prediction $\hat{y}$ by finding a set of $K$ concepts $C_1,$ $\dots, C_K$ that are masked versions of $X$. ConRAT learns concepts by maximizing the mutual information between $\bm{C}$ and $y$. We guide ConRAT to create separable and consistent concepts via two regularizers to make them human-understandable.

\subsection{Model Overview}

ConRAT is divided into three submodels: a \textbf{Concept Generator} $g_\theta (\cdot)$, which finds the concepts~$C_1, \dots, C_K$; a \textbf{Concept Selector} $s_\theta(\cdot)$, which detects whether a concept $C_k$ is present or absent (i.e., $s_k \in \{1,0\}$) in the input $X$; and a \textbf{Predictor}~$f_\theta(\cdot)$, which predicts the outcome $\hat{y}$ based on the concepts $\bm{C}$ and their presence scores $S$.

\subsubsection{Concept Generation}
\label{sec_concept_generation}Inspired by the selective rationalization field \cite{lei-etal-2016-rationalizing}, we define ``concept'' as a sequence of consecutive words in the input text. Previous studies extracted only one concept $C_1$ that is sufficient to explain the target variable $y$. In our work, a major difference is that we aim to find $K$ concepts~$C_1, \cdots, C_K$ that represent different topics or aspects and altogether explain the target variable $y$. We interpret the model as being linear in the concepts rather than depending on one overall selection of word.
More formally, we define a concept as follows:\begin{equation}
  C_k = M_k \odot X,
  \label{eq:concept}
\end{equation}where $M_k \in \mathbb{S}$ denotes a binary mask, $\mathbb{S}$ is a subset of $\mathbb{Z}_2^T$ with some constraints (introduced in Section~\ref{discovery_concepts}), and $\odot$ is the element-wise multiplication of two vectors. 

We parametrize the binary masks $\bm{M} \in \mathbb{Z}_2^{K \times T}$ with the concept generator model $g_\theta(\cdot)$, based on~a bi-directional recurrent neural network. Following previous rationalization research \cite{yu-etal-2019-rethinking,chang2020invariant}, we force $g_\theta(\cdot)$ to select one chunk of text per concept with a pre-specified length $\ell \in [1, T]$.\footnote{In early experiments, we relaxed the length constraint and generated instead $K$ differentiable masks with continuity regularizers. However, this variant produced majorly inferior results. We hypothesize that there are too many constraints to optimize with only the target label as a strong signal.}
 Instead of predicting the mask~$M_k$ directly, $g_\theta(\cdot)$ produces a score for each position $t$. Then, it samples the start position~$t^*_k$ of the chunk for each $C_k$ using the straight-through Gumbel-Softmax \cite{MaddisonMT17,JangGP17}. Finally, we compute $M_k$ as~follows:\begin{equation}
\begin{split}
  T^* &\sim Gumbel(g_\theta(X)),\\
    M_{k,t} &= \mathbbm{1}[t\in [t^*_k, min(t^*_k + \ell-1, T)]],
    \end{split} 
    \label{eq:hard_constraint}
\end{equation}
where $\mathbbm{1}$ denotes the indicator function. Although the equation is not differentiable, we can employ the straight-through technique \cite{bengio2013estimating} and approximate it with the gradient of a causal convolution and a convolution kernel of an all-one vector of length $\ell$.

\subsubsection{Concept Selection}
\label{sec_concept_selection}
A key objective of ConRAT is to produce semantically consistent and separable concepts. So far, the generator $g_\theta(\cdot)$ generates $K$ concepts for any input document. However, some documents might mention only a subset of those. Thus, the goal of the concept selector model $s_\theta(\cdot)$ is to enable ConRAT to ignore absent concepts.

Specifically, for each concept $C_k$, the model first computes a concept representation $H_{C_k}$ using a standard attention mechanism \cite{iclr2015} (the tokens whose $M_{k,t}=0$ are masked out). Then, we take the dot product of $H_{C_k}$ with a weight vector, followed by a sigmoid activation function to induce the log-probabilities of a relaxed Bernoulli distribution \cite{JangGP17}. Finally, we sample the presence score $s_k \in \{0,1\}$ of each concept independently:\begin{equation}
  S \sim RelaxedBernoulli(s_\theta(X, \bm{M})).
\end{equation}

\subsubsection{Prediction}

As inputs, the predictor $f_\theta (\cdot)$ takes the document~$X$, the masks $\bm{M}$, and the presence scores~$S$ for all concepts. First, we extract the concepts, which are masked versions of $X$. Differently than in Equation~\ref{eq:concept}, the concepts are ignored if $s_k=0$:\begin{equation}
  C_k = (M_k * s_k) \odot X.
\end{equation}
Second, the model produces the hidden representation $h'_{C_k}$ with another recurrent neural network, followed by a LeakyReLU activation function \cite{xu2015empirical}. Then, it computes the logits of $y$ by applying a linear projection for each concept: \begin{equation}
  P_k = Wh'_{C_k} + b,
\end{equation} where $W$ and $b$ are the projection parameters. Finally, $f_\theta$ computes the final outcome as follows:\begin{equation}
  p(y | C, \bm{M}, X) = softmax(\sum_{k=1}^K \alpha_k P_k s_k),
\end{equation}where $\alpha_k$ are model parameters that can be interpreted as the degree to which a particular concept contributes to the final prediction.

\subsection{Unsupervised Discovery of Concepts}
\label{discovery_concepts}

The above formulations integrate the explanation into the outcome computation. However, $M_k$ is by definition faithful to the model's inner workings but not comprehensible for the end-user. Following \citet{Alvarez-MelisJ18}, we~aim the concepts to follow three desiderata:\begin{enumerate*}
\item \textbf{Fidelity}: they should preserve relevant information,
\item \textbf{Diversity}: they should be non-overlapping and diverse, and
\item \textbf{Grounding}: they should have an immediate human-understandable interpretations
\end{enumerate*}.

The hard constraint in Equation~\ref{eq:hard_constraint} naturally enforces the grounding by forcing the concept to be a sequence of $\ell$ words. For the fidelity, it~is partly integrated in ConRAT by the prediction loss, which is the cross-entropy between the ground-truth label $y$ and the prediction $\hat{y}$: $\mathcal{L}_{pred} = CE(\hat{y},y)$. Recall that the concepts are substitutes of the input that are sufficient for the prediction. We emphasize the word ``partly'' because nothing prevents ConRAT from picking up spurious correlations.

We propose two regularizers to encourage ConRAT in finding non-overlapping, relevant, and dissimilar concepts. The first favors the orthogonality of concepts by penalizing redundant rows in $\bm{M}$:
\begin{equation}
  \mathcal{L}_{overlap} = || \bm{M}\bm{M}^T - \ell \cdot \mathbbm{1}||_F^2,
\end{equation} where $|| \cdot ||_F$ stands for the Frobenius norm of a matrix, $\mathbbm{1}$ denotes the identity matrix, and $\ell$ the pre-specified concept length. 
However, $\mathcal{L}_{overlap}$ alone does not prevent ConRAT from learning little relevant concepts. Therefore, we propose a second regularizer to encourage fidelity and diversity by minimizing the cosine similarity between the concept representations $H_{C_k}$ (see Section~\ref{sec_concept_selection}):\begin{equation}
  \mathcal{L}_{div} = \frac{1}{K}\frac{1}{K-1}\sum_{\substack{k_1,k_2=1\\k_1 \ne k_2}}^K cos(H_{C_{k_1}}, H_{C_{k_2}}).
\end{equation}

In both regularizers, we do not consider the presence scores $S$ because a model could always select only one concept; this strategy is not optimal and reduces to a special case of rationale models (i.e., $S$ would become a one-hot vector).

To summarize, the concepts are learned in an unsupervised manner and align with the three desiderata mentioned above: diversity is achieved with $\mathcal{L}_{overlap}$ and $\mathcal{L}_{div}$; fidelity is enforced by $\mathcal{L}_{pred}$ and $\mathcal{L}_{div}$, and the hard constraint in Equation~\ref{eq:hard_constraint} ensures the grounding. Finally, we train ConRAT end-to-end and minimize the loss jointly $\mathcal{\bm{L}}=\mathcal{L}_{pred} + \lambda_O \mathcal{L}_{overlap} + \lambda_D \mathcal{L}_{div}$, where $\lambda_O$ and~$\lambda_D$ control the impact of each regularizer.
\subsection{Improving  Overall Performance Further}
\label{sec_improv_perfs}
The purpose of self-explaining models is to compute outcomes while being more interpretable. However, one key point is to achieve predictive performance comparable to that of black-box models. We propose two techniques to further improve both interpretability and performance; however, ConRAT does not require these techniques to outperform other methods, as we will see later.

\paragraph{Knowledge Distillation.} We can train ConRAT not only via the information provided by the true labels but also by observing how a teacher model behaves \cite{44873}. In that case, we introduce the teacher model $T_\theta ( \cdot )$, which is a simple recurrent neural network similar to the predictor $f_\theta$. It is trained one the same data, but it uses the whole input $X$ instead of subsets selected by each~$C_k$. The overall training loss becomes $\mathcal{\bm{L}}=\mathcal{L}_{pred} + \lambda_O \mathcal{L}_{overlap} + \lambda_D \mathcal{L}_{div} + \lambda_T(\hat{y}_{T_\theta} - \hat{y}_{f_\theta})^2$.

\paragraph{Pruning Concepts.} Depending on the number of concepts and the pre-specified length, the total number of selected words can be close to or higher than the document length.\footnote{e.g., if a document contains $200$ tokens and we aim to extract $10$ concepts of $20$ tokens, all words should be selected.} In practice, it is hard to extract meaningful concepts in such settings. To alleviate this problem, we propose to prune concepts at inference and select the top-k~concepts that overlap the least with the others. More specifically, we compute the overlap as follows: for each sample in the validation set, we measure the average overlap ratio between $M_{k_1}$ and $M_{k_2}$ for each concept-pair $(C_{k_1}, C_{k_2}), k_1 \ne k_2$. Then, we select the top-k concepts whose scores are the lowest. Finally, to compute the new prediction $\hat{y}$, we update $s_k=1$ if $C_k$ is in the top-k or $s_k=0$ otherwise.

\section{Experiments}

\subsection{Datasets}
\label{datasets}
\begin{table}[t]
\centering
\begin{tabular}{@{}l@{\hspace{3mm}}c@{\hspace{3mm}}c@{}}
\multicolumn{1}{c}{\bf Dataset} & \multicolumn{1}{c}{\bf Amazon}  & \multicolumn{1}{c}{\bf Beer}\\
\toprule
\# Reviews & $24,000$ & $60,000$\\
Split Train/Val/Test & 20k/2k/2k & 50k/5k/5k\\
\# Annotations & $471$ & $994$\\
\# Human Aspects & $1$ & $5$\\
\# Words per review& $224 \pm 125$  & $184 \pm 58$\\
\end{tabular}
\caption{\label{dataset_description}Statistics of the review datasets.}
\end{table}
We evaluate the quantitative performance of ConRAT using two binary classification datasets. The first one is the single-aspect Amazon Electronics dataset \cite{ni-etal-2019-justifying}. We followed the filtering process in \citet{chang2019game} to keep only the reviews that contain evidence for both positive and negative sentiments. Specifically, we considered the first 50 tokens after the words ``pros:'' and ``cons:'' as the rationale annotations for the positive and negative labels, respectively. We randomly picked 24,000 balanced samples with ratings of four and above or two and below.

The second dataset comprises the multi-aspect beer reviews \cite{McAuley2012} used in the field of rationalization \cite{lei-etal-2016-rationalizing,yu-etal-2019-rethinking}. Each review describes various beer aspects: Appearance, Aroma, Palate, Taste, and~Overall; users also provided a five-star rating for each aspect. However, we only use the overall rating for ConRAT. The dataset includes 994 beer reviews with sentence-level aspect annotations. Following the evaluation protocol in \citet{bao-etal-2018-deriving,chang2020invariant}, we binarized the ratings $\le2$ as negative and $\ge3$ as positive. We sampled 60,000 balanced examples. Our setting is more challenging than those in previous studies because we assess the performance on all aspects (instead of three) and consider all examples for the sampling (instead of de-correlated subsets), reflecting the real data distribution. Table~\ref{dataset_description} shows the data statistics.  

\subsection{Baselines}
We consider the following baselines. \textbf{RNP} is a generator-predictor framework proposed by \citet{lei-etal-2016-rationalizing} for rationalizing neural prediction. The generator selects text spans as rationales, which are then fed to the classifier for the final prediction. \citet{yu-etal-2019-rethinking} introduced \textbf{RNP-3P}, which extends RNP to include the complement predictor as the third player. It maximizes the predictive accuracy from unselected words. The training consists of an adversarial game with the three players. \textbf{Intro-3P} \cite{yu-etal-2019-rethinking} improves RNP-3P by conditioning the generator on~the predicted outcome of a teacher model. \textbf{InvRAT} is a game-theoretic method that competitively rules out spurious words with strong correlations to the output. The game-theoretic approach \textbf{CAR} aims to infer a rationale and a counterfactual rationale that counters the true label. We follow \citet{chang2020invariant} and consider for all methods their hard constraint variant (i.e., selecting one chunk of text) with different lengths for generating~rationales.

RNP-3P and Intro-3P are trained with the policy gradient \cite{williams1992simple}. The others estimate the gradients of the rationale selections using the straight-through technique \cite{bengio2013estimating}.

All rationalization methods, except CAR, strive for a single overall selection ($K=1$) to explain the outcome. For the multi-aspect dataset, we train and tune each baseline independently for each aspect. The key difference with ConRAT is that the model is only trained on the overall aspect label and infers one rationale of $K$ concepts; the baselines are trained $K$ times to infer one rationale of one concept.
 
\subsection{Experimental Details}

To seek fair comparisons, we try to keep a similar number of parameters across all models, and we employ the same architecture for each player (generators, predictors, and discriminators/teachers) in all models: bi-directional gated recurrent units \cite{chung2014empirical} with a hidden dimension 256. We use the 100-dimensional GloVe word embeddings \cite{pennington2014glove}, Adam \cite{KingmaB14} as optimization method with a learning rate of 0.001. We set the convolutional neural network in the concept selector similarly to \cite{kim2015mind} with 3-, 5-, and 7-width filters and 50 feature maps per filter. For ConRAT, we set the regularizer factors as follow: $\lambda_O=0.05$, $\lambda_D=0.05$, and $\lambda_T=0.5$.
We use the open-source implementation for all models, and we tune them by maximizing the prediction accuracy on the dev set with $16$ random searches. For reproducibility purposes, we include additional details in Appendix~\ref{app_training}.

\begin{table}[t]
    \centering
   \caption{\label{exp_rq1_amazon}Accuracy and objective performance of rationales in automatic evaluation for the Amazon dataset.}
\begin{threeparttable}
\begin{tabular}{@{}
l@{\hspace{0.5mm}} 
c@{\hspace{1.5mm}} 
c@{\hspace{1mm}}c@{\hspace{1mm}}c@{}c@{\hspace{1.5mm}} 
c@{\hspace{1mm}}c@{\hspace{1mm}}c@{}}
& & \multicolumn{3}{c}{\textit{Factual}} & & \multicolumn{3}{c}{\textit{Counter Fact.}}\\
\cmidrule{3-5}\cmidrule{7-9}
\textbf{Model} & \textbf{Acc.} & \textbf{P} & \textbf{R} & \textbf{F} & &  \textbf{P} & \textbf{R} & \textbf{F}\\
\toprule
RNP & $\mathbf{75.5}$ &$32.6$ & $18.8$ & $23.8$ & & \multicolumn{3}{c}{$-$}\\
RNP-3P & $70.0$ & $49.4$ & $28.4$ & $36.0$ & & \multicolumn{3}{c}{$-$}\\
Intro-3P  & $75.2$ & $22.1$ & $12.8$ & $16.2$ & & \multicolumn{3}{c}{$-$}\\
InvRAT & $71.5$ & $44.3$ & $25.5$ & $32.4$ & & \multicolumn{3}{c}{$-$}\\
ConRAT-1 & $\mathbf{75.5}$ & $\mathbf{56.4}$ & $\mathbf{32.5}$ & $\mathbf{41.3}$ & & \multicolumn{3}{c}{$-$}\\
\cdashlinelr{1-9}
CAR & $73.6$& $33.0$ & $19.1$ & $24.2$ & & $\mathbf{44.1}$ & $\mathbf{25.4}$ & $\mathbf{32.2}$\\
ConRAT-6 & $\mathbf{75.4}$ & $\mathbf{50.0}$ & $\mathbf{28.8}$ & $\mathbf{36.6}$ & & $32.3$ & $18.6$ & $23.6$\\
ConRAT-4 & $75.3$ & $46.4$ & $26.7$ & $33.9$ & & $29.6$ & $17.1$ & $21.6$\\
ConRAT-2 & $75.3$ & $33.7$ & $19.4$ & $24.6$ & & $8.9$  & $5.1$  & $6.5$\\
\end{tabular}
\end{threeparttable}
\end{table}

\begin{table*}[!t]
    \centering
    \small
   \caption{\label{exp_rq2_beer_objective}Objective performance of rationales for the multi-aspect beer reviews. ConRAT only uses the overall label and ignores the other aspect labels. All baselines are trained separately on each aspect rating. \textbf{Bold} and \underline{underline} denote the best and second-best results, respectively.}
\begin{threeparttable}
\begin{tabular}{@{}c@{\hspace{1.5mm}}l@{\hspace{1.5mm}}
c@{\hspace{1.5mm}}
c@{\hspace{1mm}}c@{\hspace{1mm}}c@{}c@{\hspace{1.25mm}}
c@{\hspace{1mm}}c@{\hspace{1mm}}c@{}c@{\hspace{1.25mm}}
c@{\hspace{1mm}}c@{\hspace{1mm}}c@{}c@{\hspace{1.25mm}}
c@{\hspace{1mm}}c@{\hspace{1mm}}c@{}c@{\hspace{1.25mm}}
c@{\hspace{1mm}}c@{\hspace{1mm}}c@{}c@{\hspace{1.25mm}}
c@{\hspace{1mm}}c@{\hspace{1mm}}c@{}c@{\hspace{1.25mm}}
@{}}
& & & \multicolumn{3}{c}{\textit{Average}} & & \multicolumn{3}{c}{\textit{Appearance}} & & \multicolumn{3}{c}{\textit{Aroma}} & & \multicolumn{3}{c}{\textit{Palate}}& & \multicolumn{3}{c}{\textit{Taste}}& & \multicolumn{3}{c}{\textit{Overall}}\\
\cmidrule{4-6}\cmidrule{8-10}\cmidrule{12-14}\cmidrule{16-18}\cmidrule{20-22}\cmidrule{24-26}
& \textbf{Model} & \textbf{Acc.} & \textbf{P} & \textbf{R} &  \textbf{F} & & \textbf{P} & \textbf{R} & \textbf{F}  & & \textbf{P} & \textbf{R} & \textbf{F} & &  \textbf{P} & \textbf{R} & \textbf{F} & &  \textbf{P} & \textbf{R} & \textbf{F} & &  \textbf{P} & \textbf{R} & \textbf{F}\\
\toprule
\multirow{5}{*}{\rotatebox{90}{\textit{$\ell=20$}}}
& RNP & $81.1$ & $30.7$ & $22.1$ & $24.9$ & & $30.8$ & $23.2$ & $26.5$ & & $22.1$ & $21.0$ & $21.5$ & & $17.7$ & $24.1$ & $20.4$ & & $28.1$ & $16.7$ & $20.9$ & & $54.9$ & $25.8$ & $35.1$\\
& RNP-3P & $80.5$ & $29.1$ & $22.5$ & $25.0$ & & $30.4$ & $25.6$ & $27.8$ & & $19.3$ & $20.4$ & $19.8$ & & $10.3$ & $12.0$ & $11.1$ & & $43.9$ & $\underline{28.4}$ & $\underline{34.5}$ & & $41.6$ & $26.0$ & $32.0$\\
& Intro-3P & $\underline{85.6}$ & $24.2$ & $19.6$ & $21.3$ & & $28.7$ & $24.8$ & $26.6$ & & $14.3$ & $14.4$ & $14.3$ & & $16.6$ & $19.3$ & $17.9$ & & $24.2$ & $13.6$ & $17.4$ & & $37.0$ & $25.9$ & $30.5$\\
& InvRAT & $82.9$ & $\underline{41.8}$ & $\underline{31.1}$ & $\underline{34.8}$ & & $\underline{54.5}$ & $\underline{45.5}$ & $\underline{49.6}$ & & $\underline{26.1}$ & $\underline{27.6}$ & $\underline{26.9}$ & & $\underline{22.6}$ & $\underline{25.9}$ & $\underline{24.1}$ & & $\underline{46.6}$ & $27.4$ & $34.5$ & & $\underline{59.0}$ & $\underline{29.3}$ & $\underline{39.2}$\\
& ConRAT\tnote{*} & $\mathbf{91.4}$ & $\mathbf{50.0}$ & $\mathbf{42.0}$ & $\mathbf{44.9}$ & & $\mathbf{57.8}$ & $\mathbf{53.0}$ & $\mathbf{55.3}$ & &$\mathbf{31.9}$ & $\mathbf{35.5}$ & $\mathbf{33.6}$ & & $\mathbf{29.0}$ & $\mathbf{36.3}$ & $\mathbf{32.3}$ & & $\mathbf{56.5}$ & $\mathbf{33.9}$ & $\mathbf{42.4}$ & & $\mathbf{74.9}$ & $\mathbf{51.0}$ & $\mathbf{60.7}$\\
\midrule
\multirow{5}{*}{\rotatebox{90}{\textit{$\ell=10$}}}
& RNP & $\underline{84.4}$ & $41.3$ & $16.6$ & $23.2$ & & $40.1$ & $12.0$ & $18.5$ & & $\mathbf{33.3}$ & $\mathbf{18.7}$ & $\mathbf{24.0}$ & & $\mathbf{25.1}$ & $\mathbf{17.4}$ & $\mathbf{20.6}$ & & $32.3$ & $9.8$ & $15.07$ & & $76.0$ & $25.1$ & $37.8$\\
& RNP-3P & $83.1$ & $31.1$ & $13.5$ & $18.6$ & & $41.8$ & $19.2$ & $26.3$ & & $22.2$ & $12.4$ & $15.9$ & & $16.5$ & $10.4$ & $12.7$ & & $33.2$ & $10.6$ & $16.1$ & & $41.9$ & $14.7$ & $21.8$\\
& Intro-3P & $80.9$ & $21.8$ & $10.8$ & $14.3$ & & $51.0$ & $26.0$ & $34.4$ & & $18.8$ & $9.7$ & $12.8$ & & $16.5$ & $10.6$ & $12.9$ & & $9.7$ & $2.6$ & $4.1$ & & $13.1$ & $5.2$ & $7.4$\\
& InvRAT & $81.9$ & $\underline{47.1}$ & $\underline{17.8}$ & $\underline{25.5}$ & & $\mathbf{59.4}$ & $\underline{26.1}$ & $\mathbf{36.3}$ & & $31.3$ & $15.5$ & $20.8$ & & $16.4$ & $9.6$ & $12.1$ & & $\underline{39.1}$ & $\underline{11.6}$ & $\underline{17.9}$ & & $\mathbf{89.1}$ & $\underline{26.4}$ & $\underline{40.7}$\\
& ConRAT\tnote{*} & $\mathbf{91.3}$ & $\mathbf{48.1}$ & $\mathbf{20.1}$ & $\mathbf{28.0}$ & & $\underline{51.7}$ & $\mathbf{26.2}$ & $\underline{34.8}$ & & $\underline{32.6}$ & $\underline{17.4}$ & $\underline{22.7}$ & & $\underline{23.0}$ & $\underline{13.8}$ & $\underline{17.3}$ & & $\mathbf{45.3}$ & $\mathbf{13.1}$ & $\mathbf{20.3}$ & & $\underline{88.0}$ & $\mathbf{30.1}$ & $\mathbf{44.9}$\\
\end{tabular}
\begin{tablenotes}
     \item[*]\small The model is only trained on the overall label and does not have access to the other ground-truth labels.
  \end{tablenotes}
\end{threeparttable}
\end{table*}

\subsection{RQ 1: Can ConRAT find evidence for factual and counterfactual rationales?}
\label{rq1}
We aim to validate whether ConRAT can identify the two evidences for positive and negative sentiments. We set the concept length $\ell=30$, we compare the generated rationales with the annotations, and we report the precision, recall, and F1 score. In this experiment, no teacher is used in ConRAT.

Table~\ref{exp_rq1_amazon} contains the results. The top rows contain the results when only the factual rationales are considered for the evaluation, and ConRAT-1 uses only one concept. We see that ConRAT surpasses the baselines in finding rationales that align with human annotations, and it also matches the test accuracy with the baselines. Interestingly, we note that the baselines achieving the highest accuracy underperform in finding the correct rationales.

For the factual and counterfactual rationales, CAR finds one rationale to support the outcome and another one to counter it, in an adversarial game. However, the concepts inferred by ConRAT are not guaranteed to align with the rationales~as there is no explicit signal to infer counterfactual concepts. Thus, we increase the number of concepts up to six and prune ConRAT to consider only the two most dissimilar concepts (see Section~\ref{sec_improv_perfs}).

The bottom of Table~\ref{exp_rq1_amazon} show the results. With only two concepts, ConRAT-2 outperforms CAR in terms of test accuracy and  matches~the performance for the factual rationales, but it poorly identifies counterfactual rationales. However, there is~a major improvement when we increase the number of concepts and use pruning. Indeed, the word~distribution of the factual and counterfactual rationales are different, hence captured with pruning. ConRAT's factual rationales are better than those of all models. The counterfactual ones~get~closer to those produced by CAR. We show later~in Section~\ref{rq3} that pruning helps in achieving better correlation with human judgments but is not required.

\begin{figure*}[!t]
\centering
\begin{tabular}{@{}c@{\hspace{1mm}}c@{\hspace{1mm}}c@{}}
  ConRAT (Ours) & InvRAT \cite{chang2020invariant} & RNP \cite{lei-etal-2016-rationalizing}\\
     \includegraphics[width=0.35\textwidth,height=2.3cm]{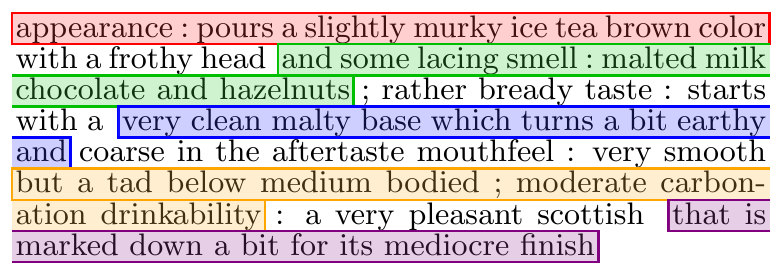} &
     \includegraphics[width=0.35\textwidth,height=2.3cm]{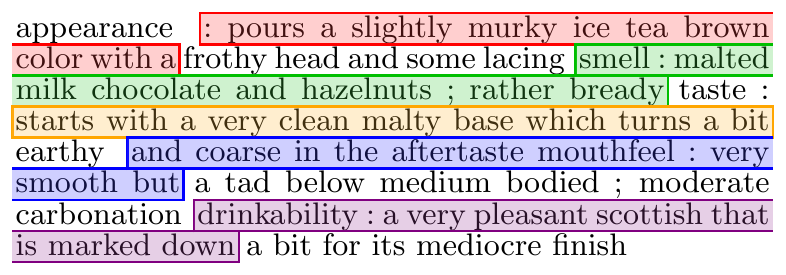} &
     \includegraphics[width=0.35\textwidth,height=2.3cm]{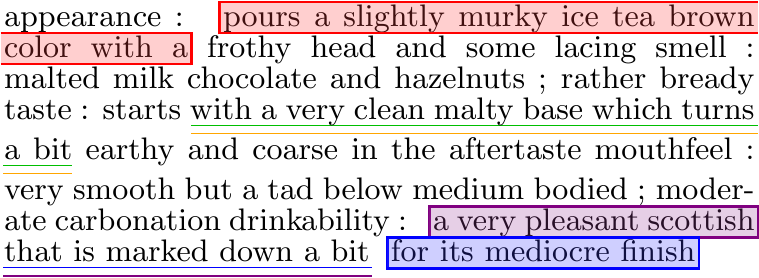} 
\end{tabular}
\caption{\label{sample}Concepts generated (with $\ell$=10)~for~a beer review. \underline{Underline} highlights ambiguities. The color depicts the aspects: \MyDBox[red]{\strut Appearance}, \MyDBox[green]{\strut Aroma}, \MyDBox[orange]{\strut Palate}, \MyDBox[blue]{\strut Taste}, and~\MyDBox[purple]{\strut Overall} . \textbf{ConRAT is trained only on the overall label}.} 
\end{figure*}

\subsection{RQ 2: Are concepts inferred by ConRAT consistent with human rationalization?}

We investigate whether ConRAT can recover all beer aspects by using only the overall ratings. Because beer reviews are smaller in length than Amazon ones, we set the concept length $\ell$ to 10 and 20. We fix the number of concepts to ten and prune ConRAT to keep five. We manually map them to the closest aspect for comparison. We trained the teacher model, used in Intro-3P and ConRAT, and obtained 91.4\% accuracy. More results and illustrations are available in Appendix~\ref{app_rq2} and \ref{app_samples}.

\paragraph{Objective Evaluation.}
\label{rq2_exp_obj}
Similar to Section~\ref{rq1}, we compare the generated rationales with the human annotations on the five aspects and the average performance. The main results are shown in Table~\ref{exp_rq2_beer_objective}. On average, ConRAT achieves the best performance while trained only on the overall ratings. This shows that the generated concepts, learned in an unsupervised manner, are separable, consistent, and correlated with human judgments to a certain extent. For the concept length $\ell=20$, ConRAT produces significant superior results for all aspects, whereas the difference with InvRAT is less pronounced for $\ell=10$. Finally, ConRAT's concepts lead to the highest accuracy and respect the grounding desideratum, thanks to the teacher.

We hypothesize that the baselines underperform due to the high correlations among the aspect ratings. Thus, they are more prone to pick up spurious correlations between the input features and the output. By considering multiple concepts simultaneously, ConRAT reduces the impact of spurious correlations. Regarding Intro-3P and RNP-3P, both suffer from instability issues due to the policy gradient \cite{chang2020invariant,yu-etal-2019-rethinking}.

We visualize an example in Figure~\ref{sample}. We observe that ConRAT induces interpretable~concepts, while the best baselines suffer from spurious correlations. By reading our concepts alone, humans will easily predict the aspect label and its polarity.

\paragraph{Subjective Evaluation.}
\label{rq2_exp_subj}
We conduct a human evaluation using Amazon’s Mechanical Turk (details in Appendix~\ref{app_rq2_hum}) to judge the understandability of the concepts. Following \citet{chang2019game}, we sampled 100 balanced reviews from the hold-out set for each aspect, model, and concept length, resulting in 5,000 samples. We showed the examples in random order. An evaluator is presented with the concept generated by one of the five methods (unselected words are not visible). We credit a success when the evaluator guesses the true aspect label and its sentiment. We report the success rate as the performance metric. A random guess has a 10\% success rate.
\begin{figure}[!t]
\centering
\includegraphics[width=0.469\textwidth]{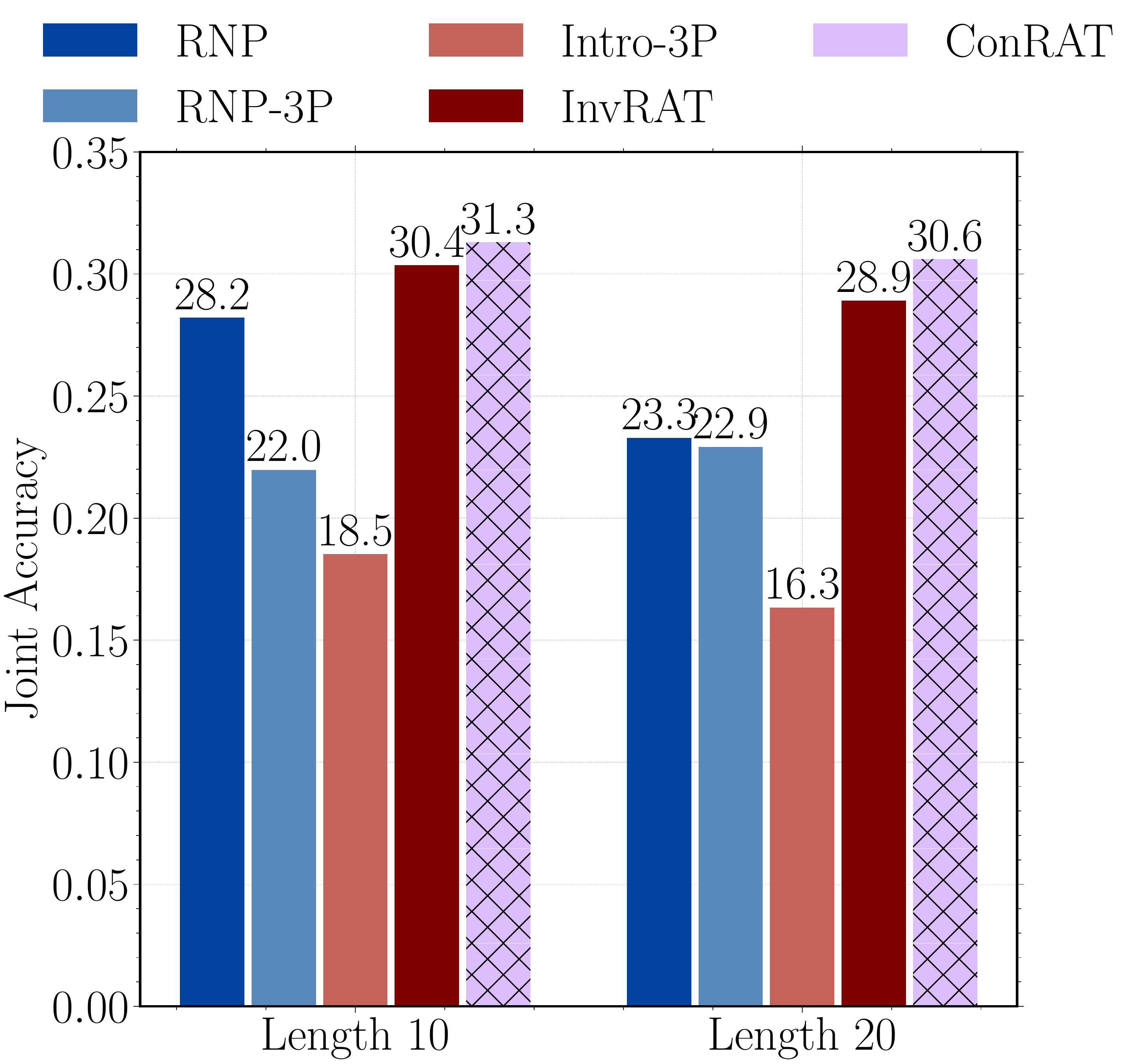}
\caption{\label{exp_rq2_beer_sub}Subjective performance of rationales for the multi-aspect beer reviews. Evaluators need to guess both the sentiment and what aspect the concept is about, which makes random guess only 10\%.}
\end{figure}

Figure~\ref{exp_rq2_beer_sub} shows the main results. Similar to the objective evaluation, ConRAT reaches the best performance, followed by InvRAT. Moreover, ConRAT only requires a single training on the overall aspect. It emphasizes that the discovered concepts satisfy the fidelity and diversity desiderata and better correlate with human judgments compared with supervised baselines.

\subsection{RQ 3: How does the number of concepts $K$ in ConRAT affect the performance?}
\label{rq3}

We study the impact of the number of concepts $K$ in ConRAT on the performance, as discussed in Section~\ref{rq2_exp_obj}. We set the number of concepts to~the number of aspects ($K$=5) and then increase it to $K$=10 and $K$=20. We prune ConRAT to keep only the five most dissimilar concepts (see Section~\ref{sec_improv_perfs}). 

Results are shown in Table~\ref{exp_rq3}. First, we observe that the performance is already better than the baselines in Table~\ref{exp_rq2_beer_objective} with $K$=5. Second, when increasing $K$ and pruning ConRAT, the performance is boosted further. However, we remark that the interpretability of the concepts follows a bell curve and significantly decreases when $K$=20. One potential reason is that we expect overlaps between the discriminative concepts that relate to beer aspects.\footnote{As shown in Table~\ref{dataset_description}, the mean length of beer reviews is 184 words. With $\ell$=20 and $C$=20, 400 words are highlighted.}
Thus, the five most dissimilar concepts might align less with human-defined concepts.

\begin{table}[t]
    \centering
   \caption{\label{exp_rq3}Impact of the number of concepts in ConRAT on the objective performance for the beer reviews.}
\begin{threeparttable}
\begin{tabular}{@{}
cc
cccc@{}}
& & & \multicolumn{3}{c}{\textit{Average}} \\
\cmidrule{4-6}
\multicolumn{2}{c}{\textbf{\#Concepts}} & \textbf{Acc.} & \textbf{P} & \textbf{R} & \textbf{F}\\
\toprule
\multirow{3}{*}{\rotatebox{90}{\textit{$\ell=20$}}}
& $K=5\ \ $ & $90.95$ & $48.96$ & $37.59$ & $41.37$\\
& $K=10$ & $\mathbf{91.35}$ & $\mathbf{50.02}$ & $\mathbf{41.96}$ & $\mathbf{44.86}$\\
& $K=20$ & $90.24$ & $37.78$ & $31.19$ & $32.84$\\
\midrule
\multirow{3}{*}{\rotatebox{90}{\textit{$\ell=10$}}}
& $K=5\ \ $ & $89.64$ & $47.60$ & $19.23$ & $26.90$\\
& $K=10$ & $\mathbf{91.25}$ & $\mathbf{48.12}$ & $\mathbf{20.11}$ & $\mathbf{27.97}$\\
& $K=20$ & $91.05$ & $35.71$ & $14.84$ & $20.71$\\
\end{tabular}
\end{threeparttable}
\end{table}

\subsection{RQ 4: How does each module of ConRAT contribute to the overall performance?}

Finally, we analyze the importance of each module in an ablation study. To avoid any bias from pruning, we set the number of concepts to five.\footnote{We obtain similar results with $K$=10 and $K$=20.}

Table~\ref{exp_rq4} shows the results. When ConRAT ignores the overlapping or the diversity regularizer, we observe a large drop in the rationale performance. This is expected as the diversity desideratum is not encouraged anymore. However, we remark that the sentiment prediction accuracy increases, which is certainly caused by spurious correlation with the ground-truth label.
When all concepts are considered ($s_k=1$ $\forall k$), we note that the sentiment accuracy stays similar. However, the objective performance decreases by 10\% for the precision and more than 20\% for the recall and F1 score. These results align with prior work: users write opinions about the topics they care about \cite{musat2015personalizing,antognini2020interacting}. ConRAT reduces the noise at training by selecting concepts described in the current document. Finally, the teacher model helps ConRAT to boost the sentiment accuracy by more than 3\% absolute score, without affecting the rationale quality.

\begin{table}[t]
    \centering
   \caption{\label{exp_rq4}Ablation study of ConRAT with five concepts.}
\begin{threeparttable}
\begin{tabular}{@{}
l@{\hspace{1mm}}
c@{\hspace{2.5mm}}c@{\hspace{2.5mm}}c@{\hspace{2.5mm}}c@{}}
& & \multicolumn{3}{c}{\textit{Average}} \\
\cmidrule{3-5}
\textbf{Model}& \textbf{Acc.} & \textbf{P} & \textbf{R} & \textbf{F}\\
\toprule
ConRAT  & $89.64$ & $47.60$ & $19.23$ & $26.90$\\
\ \ - No $\mathcal{L}_{overlap}$ & $91.05$ & $31.50$ & $13.16$ & $18.37$\\
\ \ - No $\mathcal{L}_{div}$ & $90.85$ & $34.49$ & $11.69$ & $16.95$\\
\ \ - No $s_\theta(\cdot):$$s_k=1 \forall k$ & $89.74$ & $43.13$ & $14.95$ & $21.42$\\
\ \ - No Teacher & $86.52$ & $45.31$ & $19.65$ & $26.99$\\
\end{tabular}
\end{threeparttable}
\end{table}

\section{Conclusion}

Providing explanations for automated predictions carries much more impact, increases transparency, and might even be vital. Previous works have~proposed using rationale methods to explain the prediction of a target variable. However, they do not properly capture the multi-faceted nature of useful rationales. We proposed ConRAT, a novel self-explaining model that extracts a set of concepts and explains the outcome with a linear aggregation of concepts, similar to how humans~reason.

Our second contribution is two novel regularizers that guide ConRAT to generate interpretable concepts. Experiments on both single- and multi-aspect sentiment classification datasets show that ConRAT, by using only the overall label, is the first to provide superior rationale and predictive performance compared with supervised state-of-the-art methods trained for each aspect label. Moreover, ConRAT produces concepts considered superior in interpretability when evaluated by humans.

\bibliographystyle{acl_natbib}

\clearpage
\appendix

\section{Additional Training Details}
\label{app_training}

We tune all models on the dev set. We truncate all reviews to $320$ tokens for the beer dataset and $400$ tokens for Amazon reviews. We have operated a random search over $16$ trials. All baselines, except CAR, are tuned for each aspect ($80$ trials in total for the five aspects). We chose the models achieving the lowest validation accuracy. Most of the time, all models converged under $30$ epochs. The range of hyperparameters are the following for ConRAT (similar for other models):

\begin{itemize}
  \item Learning rate: $[0.0005, 0.00075, 0.001]$;
  \item Batch size: $[128]$;
  \item Hidden size: $[256]$;
  \item $\lambda_D$: $[0.01, 0.05, 0.1, 0.25, 0.5, 0.75, 1.0]$;
  \item $\lambda_O$: $[0.01, 0.05, 0.1, 0.25, 0.5, 0.75, 1.0]$;
  \item $\lambda_T$: $[0.5, 0.6]$;
  \item Dropout: $[0.0, 0.1]$;
  \item Weight decay: $[0.0, 10^{-8}, 10^{-10}]$;
  \item Gumbel temperature in $f_\theta(\cdot)$: $[1.0;1.5]$;
  \item Gumbel temperature in $s_\theta(\cdot)$: $[1.0;1.5]$;
\end{itemize}

\subsection{Hardware / Software}

\begin{itemize}
  \item \textbf{CPU}: 2x Intel Xeon E5-2680 v3, 2x 12 cores, 24 threads, 2.5 GHz, 30 MB cache;
  \item \textbf{RAM}: 16x16GB DDR4-2133;
  \item \textbf{GPU}: 2x Nvidia Titan X Maxwell;
  \item \textbf{OS}: Ubuntu 18.04;
  \item \textbf{Software}: Python 3, PyTorch 1.3, CUDA 10.
\end{itemize}

\section{Complementary Results RQ 2}
\label{app_rq2}

\subsection{Objective Evaluation}
\label{app_rq2_obj}

The results for the concept length $\ell=5$ is shown in Table~\ref{exp_rq2_beer_objective_5}. 

Moreover, we report in Table~\ref{app_rq2_beer_sent} the performance for the \textbf{unsupervised} sentiment prediction task for the aspects whose labels are not available to ConRAT: Appearance, Aroma, Palate, and Taste. As we can see, ConRAT achieves competitive results compared to \textbf{supervised} baselines.

\begin{table*}[t]
    \centering
    \small
   \caption{\label{exp_rq2_beer_objective_5}Objective performance of rationales for the multi-aspect beer reviews with the concept length set to five. ConRAT only uses the overall rating and does not have access to the other aspect labels. All baselines are trained separately on each aspect label. \textbf{Bold} and \underline{underline} denote the best and second-best results, respectively.}
\begin{threeparttable}
\begin{tabular}{@{}c@{\hspace{1.5mm}}l@{\hspace{1.5mm}}
c@{\hspace{1.5mm}}
c@{\hspace{1mm}}c@{\hspace{1mm}}c@{}c@{\hspace{1.25mm}}
c@{\hspace{1mm}}c@{\hspace{1mm}}c@{}c@{\hspace{1.25mm}}
c@{\hspace{1mm}}c@{\hspace{1mm}}c@{}c@{\hspace{1.25mm}}
c@{\hspace{1mm}}c@{\hspace{1mm}}c@{}c@{\hspace{1.25mm}}
c@{\hspace{1mm}}c@{\hspace{1mm}}c@{}c@{\hspace{1.25mm}}
c@{\hspace{1mm}}c@{\hspace{1mm}}c@{}c@{\hspace{1.25mm}}
@{}}
& & & \multicolumn{3}{c}{\textit{Average}} & & \multicolumn{3}{c}{\textit{Appearance}} & & \multicolumn{3}{c}{\textit{Aroma}} & & \multicolumn{3}{c}{\textit{Palate}}& & \multicolumn{3}{c}{\textit{Taste}}& & \multicolumn{3}{c}{\textit{Overall}}\\
\cmidrule{4-6}\cmidrule{8-10}\cmidrule{12-14}\cmidrule{16-18}\cmidrule{20-22}\cmidrule{24-26}
& \textbf{Model} & \textbf{Acc.} & \textbf{P} & \textbf{R} &  \textbf{F} & & \textbf{P} & \textbf{R} & \textbf{F}  & & \textbf{P} & \textbf{R} & \textbf{F} & &  \textbf{P} & \textbf{R} & \textbf{F} & &  \textbf{P} & \textbf{R} & \textbf{F} & &  \textbf{P} & \textbf{R} & \textbf{F}\\
\toprule
\multirow{5}{*}{\rotatebox{90}{\textit{$\ell=5$}}}
& RNP & $80.8$ & $41.3$ & $10.4$ & $16.4$ & & $\underline{50.9}$ & $\mathbf{13.3}$ & $\mathbf{21.1}$ & & $\mathbf{43.2}$ & $\mathbf{12.7}$ & $\mathbf{19.7}$ & & $\underline{27.1}$ & $\underline{10.0}$ & $\underline{14.5}$ & & $5.5$ & $0.59$ & $1.07$ & & $\underline{80.0}$ & $\underline{15.3}$ & $\underline{25.7}$\\
& RNP-3P & $81.5$ & $32.9$ & $6.9$ & $11.2$ & & $35.1$ & $7.3$ & $12.1$ & & $25.6$ & $7.2$ & $11.3$ & & $17.0$ & $5.2$ & $8.0$ & & $28.6$ & $4.0$ & $7.1$ & & $58.2$ & $10.5$ & $17.8$\\
& Intro-3P & $\underline{84.6}$ & $29.8$ & $7.0$ & $11.3$ & & $47.3$ & $12.4$ & $19.7$ & & $35.4$ & $9.9$ & $15.5$ & & $9.7$ & $2.8$ & $4.3$ & & $24.3$ & $3.8$ & $6.6$ & & $32.4$ & $6.3$ & $10.6$\\
& InvRAT & $83.6$ & $\underline{46.4}$ & $\mathbf{11.4}$ & $\mathbf{18.1}$ & & $\mathbf{51.0}$ & $\underline{13.1}$ & $\underline{20.8}$ & & $\underline{40.6}$ & $\underline{11.9}$ & $\underline{18.4}$ & & $\mathbf{32.0}$ & $\mathbf{11.8}$ & $\mathbf{17.2}$ & & $\underline{36.1}$ & $\underline{5.6}$ & $\underline{9.6}$ & & $72.5$ & $14.7$ & $24.4$\\
& ConRAT\tnote{\textdagger} & $\mathbf{90.4}$ &$\mathbf{46.6}$ & $\underline{10.9}$ & $\underline{17.5}$ & & $47.2$ & $12.4$ & $19.6$ & & $26.9$ & $7.1$ & $11.3$ & & $26.6$ & $9.2$ & $13.7$ & & $\mathbf{39.2}$ & $\mathbf{6.2}$ & $\mathbf{10.8}$ & & $\mathbf{93.1}$ & $\mathbf{19.5}$ & $\mathbf{32.21}$\\
\end{tabular}
\begin{tablenotes}
     \item[*]\small The model is only trained on the overall label and does not have access to the other ground-truth labels.
  \end{tablenotes}
\end{threeparttable}
\end{table*}

\begin{table}[t]
    \centering
   \caption{\label{app_rq2_beer_sent} Performance on the overall sentiment and the aspects whose labels are not available to ConRAT. \textbf{Bold} and \underline{underline} denote the best and second-best results.}
\begin{threeparttable}
\begin{tabular}{@{}c@{\hspace{2mm}}l@{\hspace{2mm}} c@{\hspace{2mm}}c@{\hspace{2mm}}c@{\hspace{2mm}}c@{\hspace{2mm}}c@{}}
& \textbf{Model} & \textbf{Ap.}\tnote{*} & \textbf{Ar.}\tnote{*} &  \textbf{P}\tnote{*} & \textbf{T}\tnote{*} & \textbf{O}\\
\toprule
\multirow{5}{*}{\rotatebox{90}{\textit{$\ell=5$}}}
& RNP & $\mathbf{95.98}$ & $89.74$ & $92.55$ & $79.78$ & $80.78$\\
& RNP-3P & $92.97$ & $87.11$ & $88.09$ & $73.93$ & $81.54$\\
& Intro-3P & $93.07$ & $88.38$ & $86.33$ & $77.05$ & $\underline{84.57}$\\
& InvRAT & $\mathbf{95.98}$ & $\underline{90.44}$ & $\underline{92.66}$ & $\underline{88.63}$ & $83.60$\\
& ConRAT & $91.75$\tnote{*} & $\mathbf{91.85}$\tnote{*} & $\mathbf{94.37}$\tnote{*}   & $\mathbf{92.35}$\tnote{*} & $\mathbf{90.44}$\\
\midrule
\multirow{5}{*}{\rotatebox{90}{\textit{$\ell=10$}}}
& RNP & $\underline{95.17}$ & $\mathbf{92.15}$ & $\mathbf{90.74}$ & $82.80$ & $\underline{84.41}$\\
& RNP-3P & $93.55$ & $88.48$ & $\underline{90.43}$ & $77.15$ & $83.11$\\
& Intro-3P & $93.55$ & $87.01$ & $87.21$ & $83.20$ & $80.86$\\
& InvRAT & $\mathbf{95.77}$ & $90.54$ & $89.03$ & $\underline{85.01}$ & $81.89$\\
& ConRAT & $92.25$\tnote{*} & $\underline{91.05}$\tnote{*} &  $83.80$\tnote{*}  & $\mathbf{91.85}$\tnote{*} & $\mathbf{91.25}$\\
\midrule
\multirow{5}{*}{\rotatebox{90}{\textit{$\ell=20$}}}
& RNP & $\mathbf{96.08}$ & $\mathbf{92.15}$ & $\mathbf{94.37}$ & $\mathbf{87.02}$ & $81.09$\\
& RNP-3P & $92.48$ & $\underline{87.70}$ & $89.16$ & $81.74$ & $80.47$\\
& Intro-3P & $93.46$ & $87.11$ & $88.96$ & $\underline{86.82}$ & $\underline{85.64}$\\
& InvRAT & $\underline{95.88}$ & $91.05$ & $\underline{89.44}$ & $85.11$ & $82.90$\\
& ConRAT & $67.71$\tnote{*} & $74.85$\tnote{*} &  $77.16$\tnote{*}  & $80.58$\tnote{*} & $\mathbf{91.35}$\\
\end{tabular}
\begin{tablenotes}
     \item[*]\small ConRAT predicts the sentiment of the aspect in an unsupervised fashion.
  \end{tablenotes}
\end{threeparttable}
\end{table}

\subsection{Human Evaluation Details}
\label{app_rq2_hum}

We use Amazon's Mechanical Turk crowdsourcing platform to recruit human annotators to evaluate the quality of extracted justifications and the generated justifications produced by each model. To ensure high-quality of the collected data, we restricted the pool to native English speakers from the U.S., U.K., Canada, or Australia. Additionally, we set the worker requirements at a 98\% approval rate and more than 1,000 HITS.

\begin{figure*}[t]
\centering
\includegraphics[width=0.9\textwidth,height=7cm]{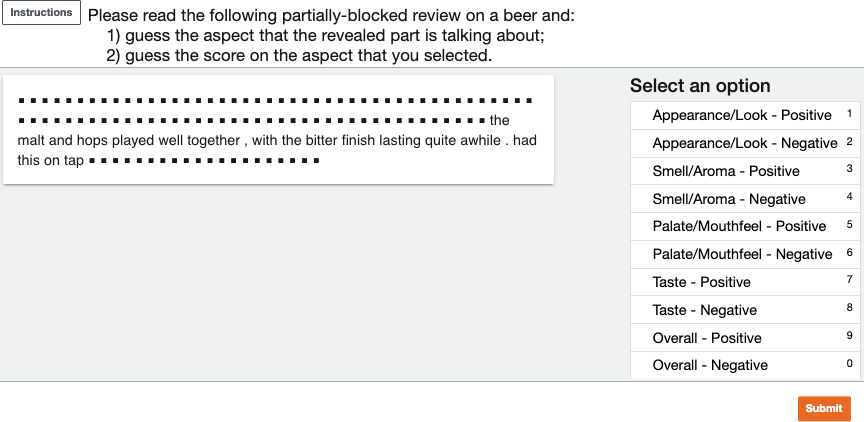}
\caption{\label{he_int_subj}Annotation platform for judging the quality of the concepts in the subjective evaluation on beer reviews.}
\end{figure*}

The user interface used to judge the quality of the justifications extracted from different methods, in Section~\ref{rq2_exp_subj}, is shown in Figure~\ref{he_int_subj}.

\subsection{Subjective Evaluation}
\label{app_rq2_sub}

\begin{figure*}[t]
\centering
\begin{subfigure}[t]{\textwidth}
\centering
\includegraphics[width=0.8\textwidth]{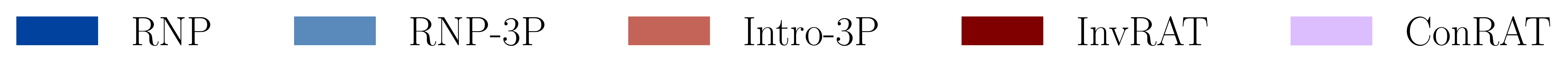}
\end{subfigure}

\begin{subfigure}[t]{0.325\textwidth}
\centering
\includegraphics[width=\textwidth]{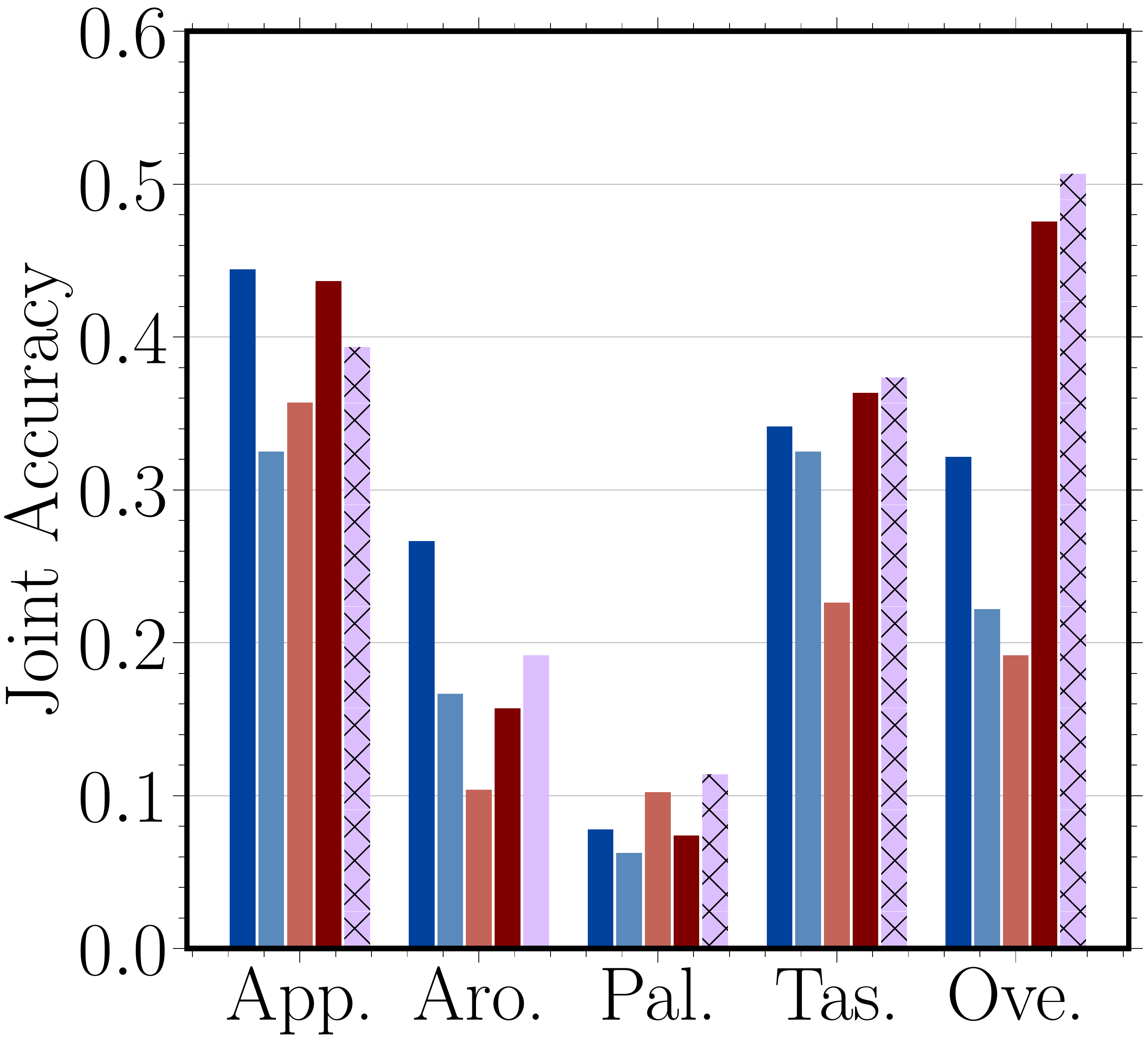}
\end{subfigure}
\begin{subfigure}[t]{0.325\textwidth}
\centering
\includegraphics[width=\textwidth]{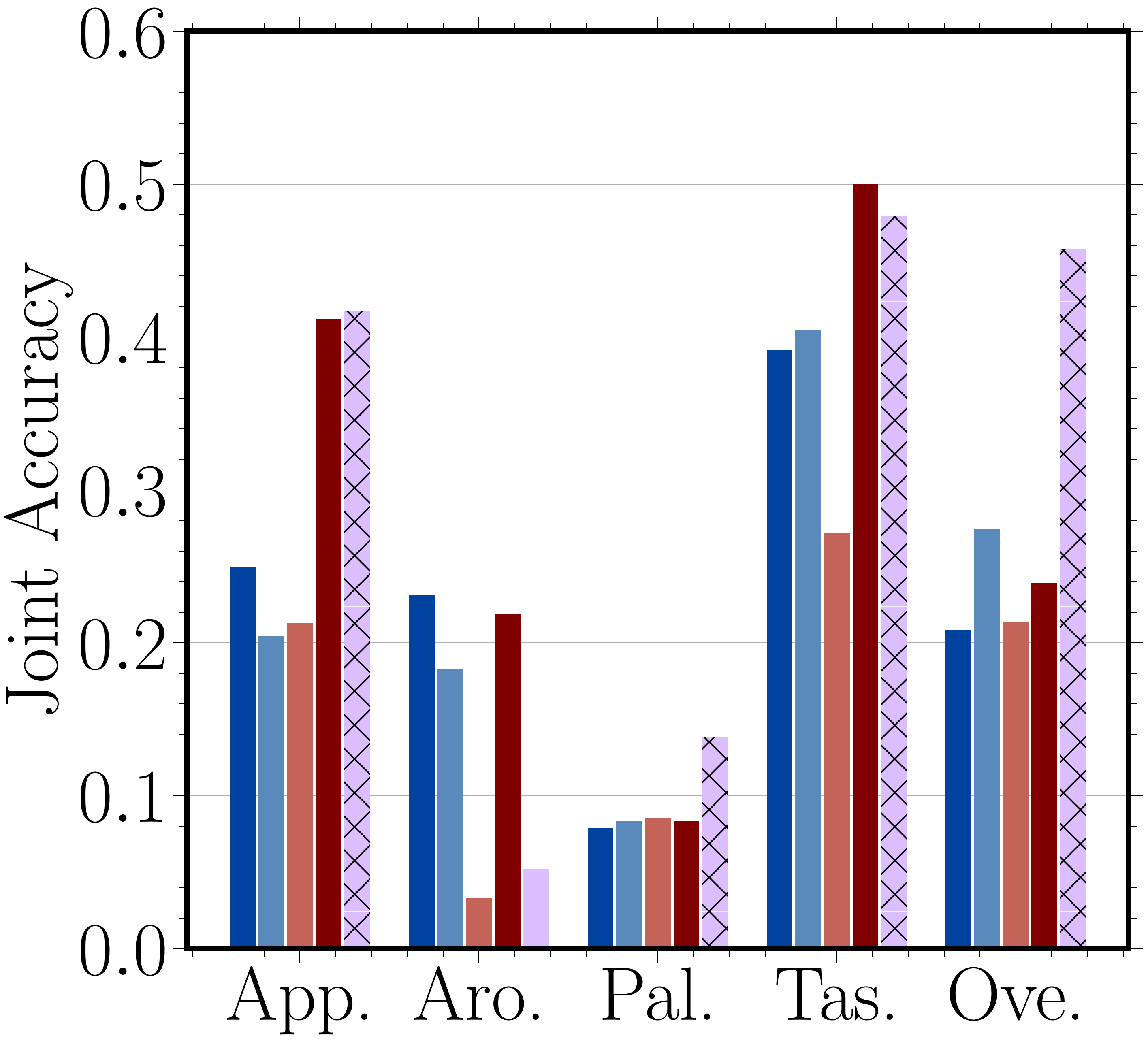}
\end{subfigure}
\begin{subfigure}[t]{0.325\textwidth}
\centering
\includegraphics[width=\textwidth,height=4.65cm]{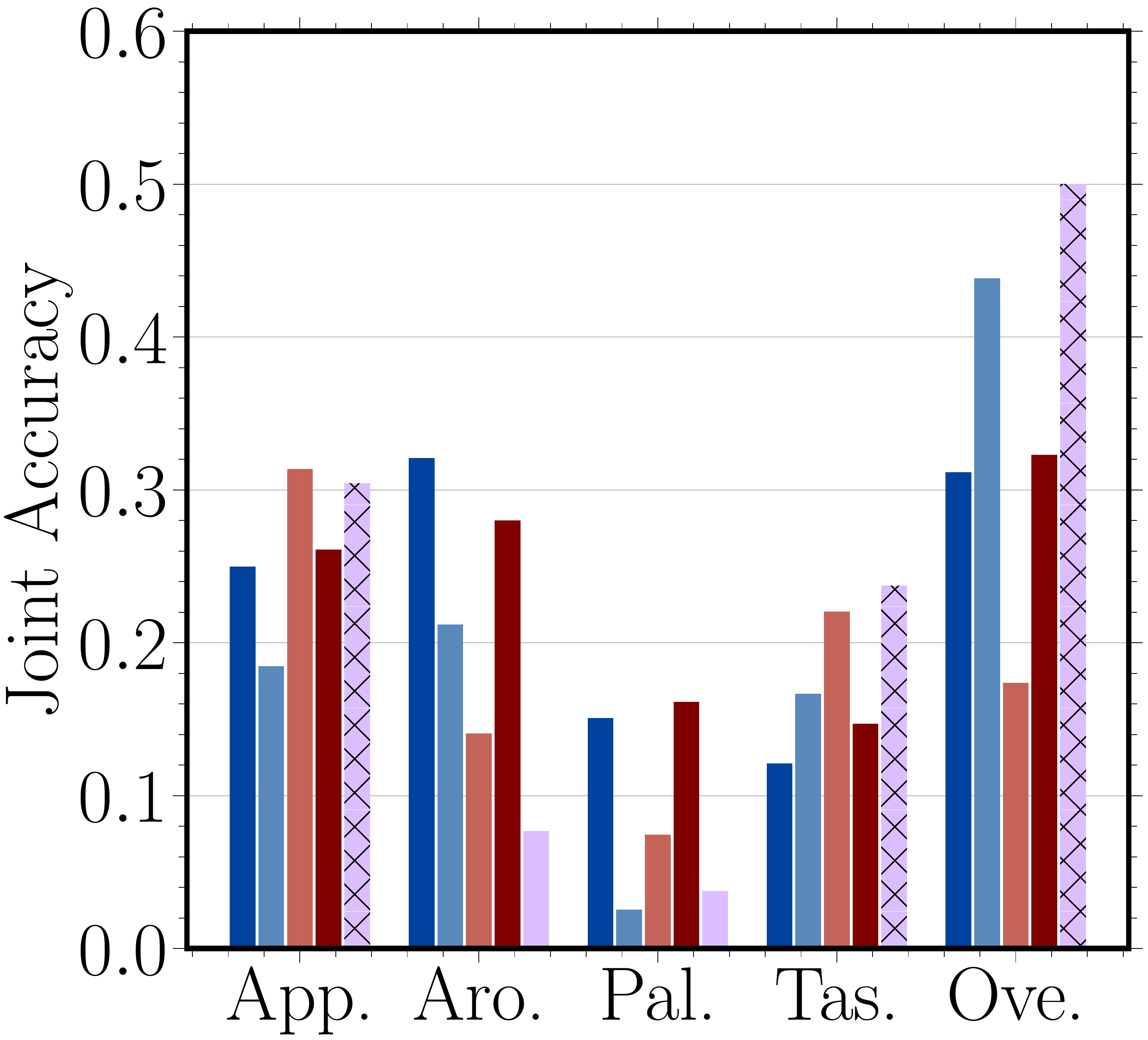}
\end{subfigure}
\par\bigskip
\par\bigskip

\begin{subfigure}[t]{0.325\textwidth}
\centering
\includegraphics[width=\textwidth]{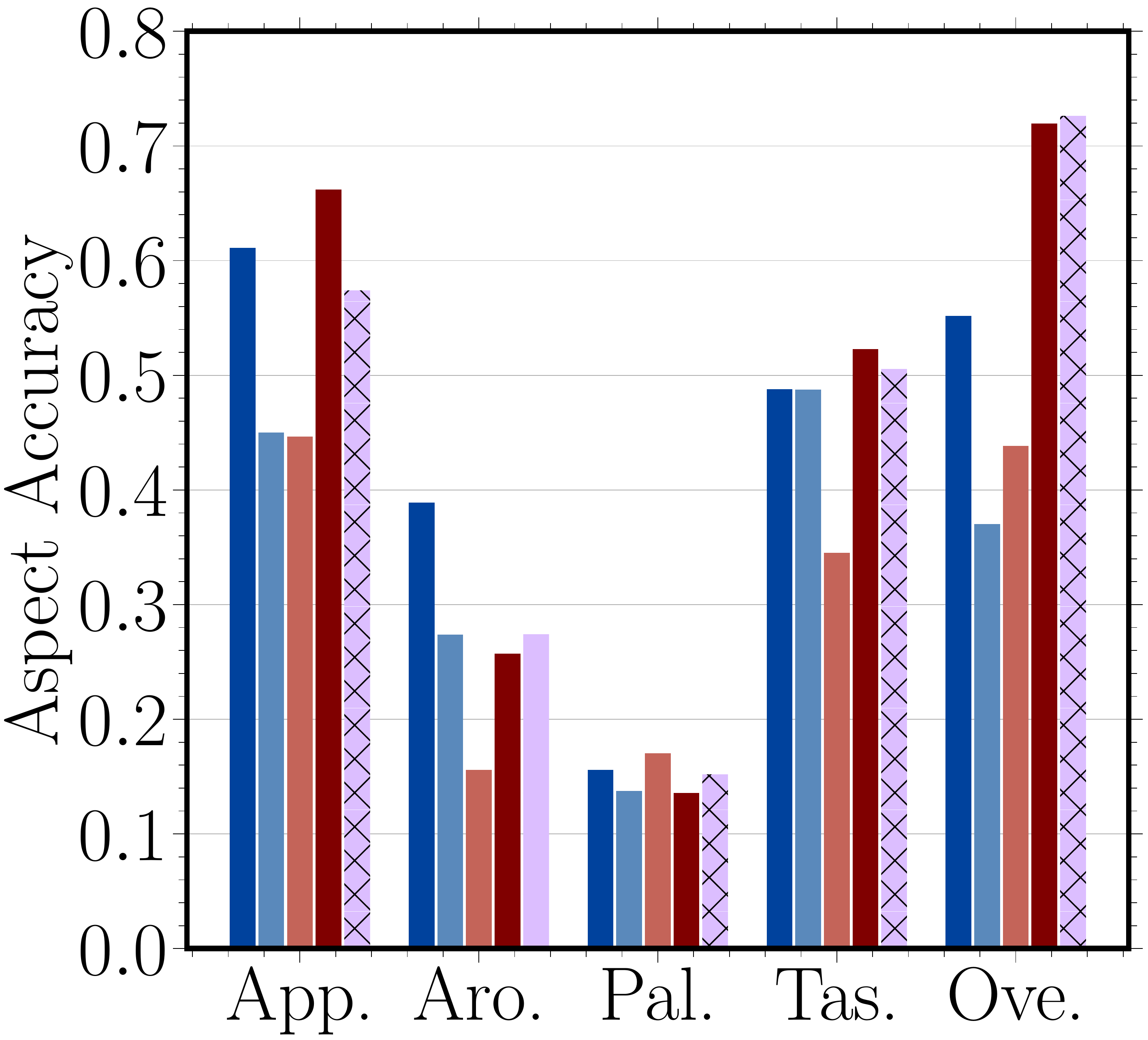}
\end{subfigure}
\begin{subfigure}[t]{0.325\textwidth}
\centering
\includegraphics[width=\textwidth]{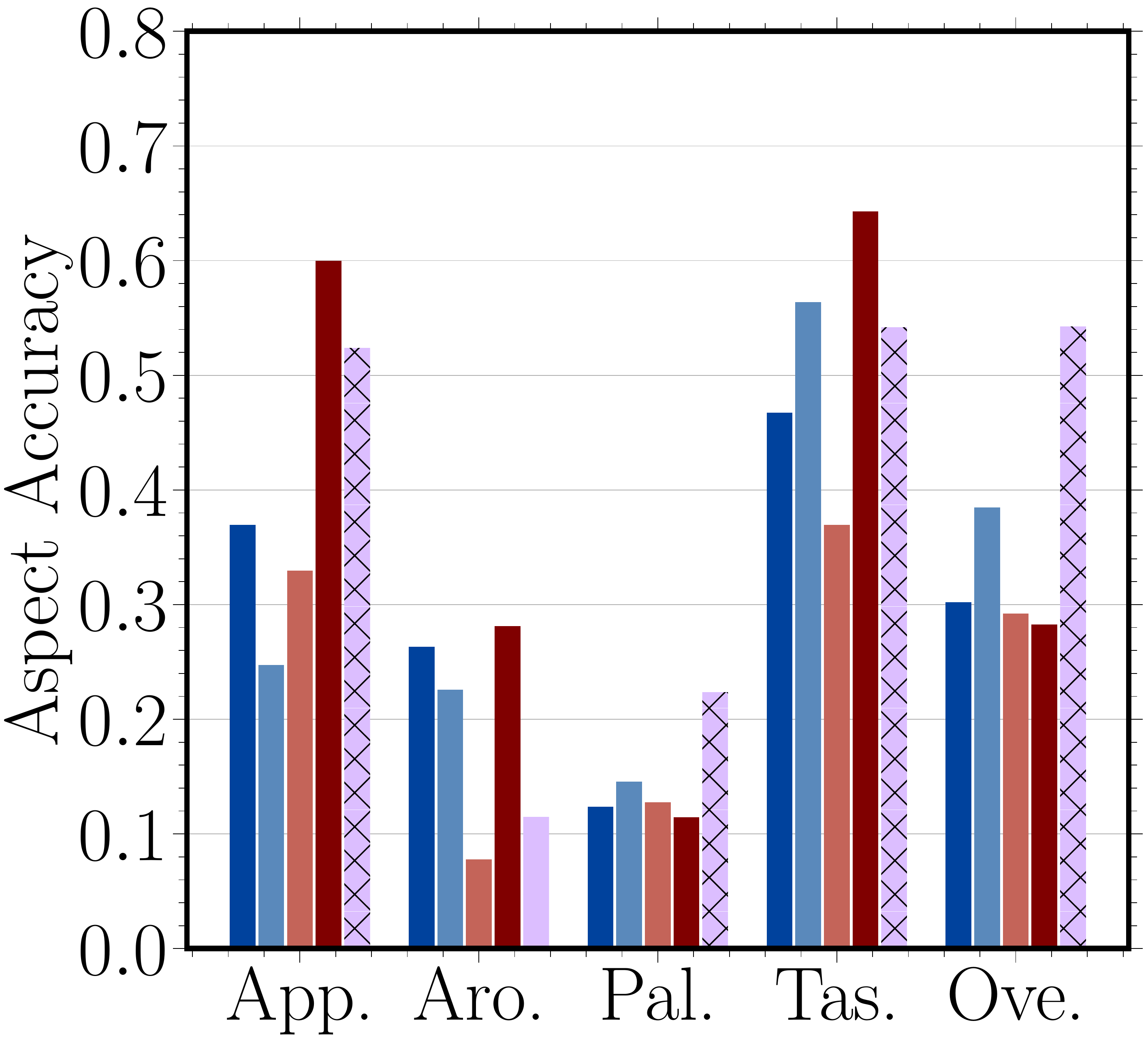}
\end{subfigure}
\begin{subfigure}[t]{0.325\textwidth}
\centering
\includegraphics[width=\textwidth]{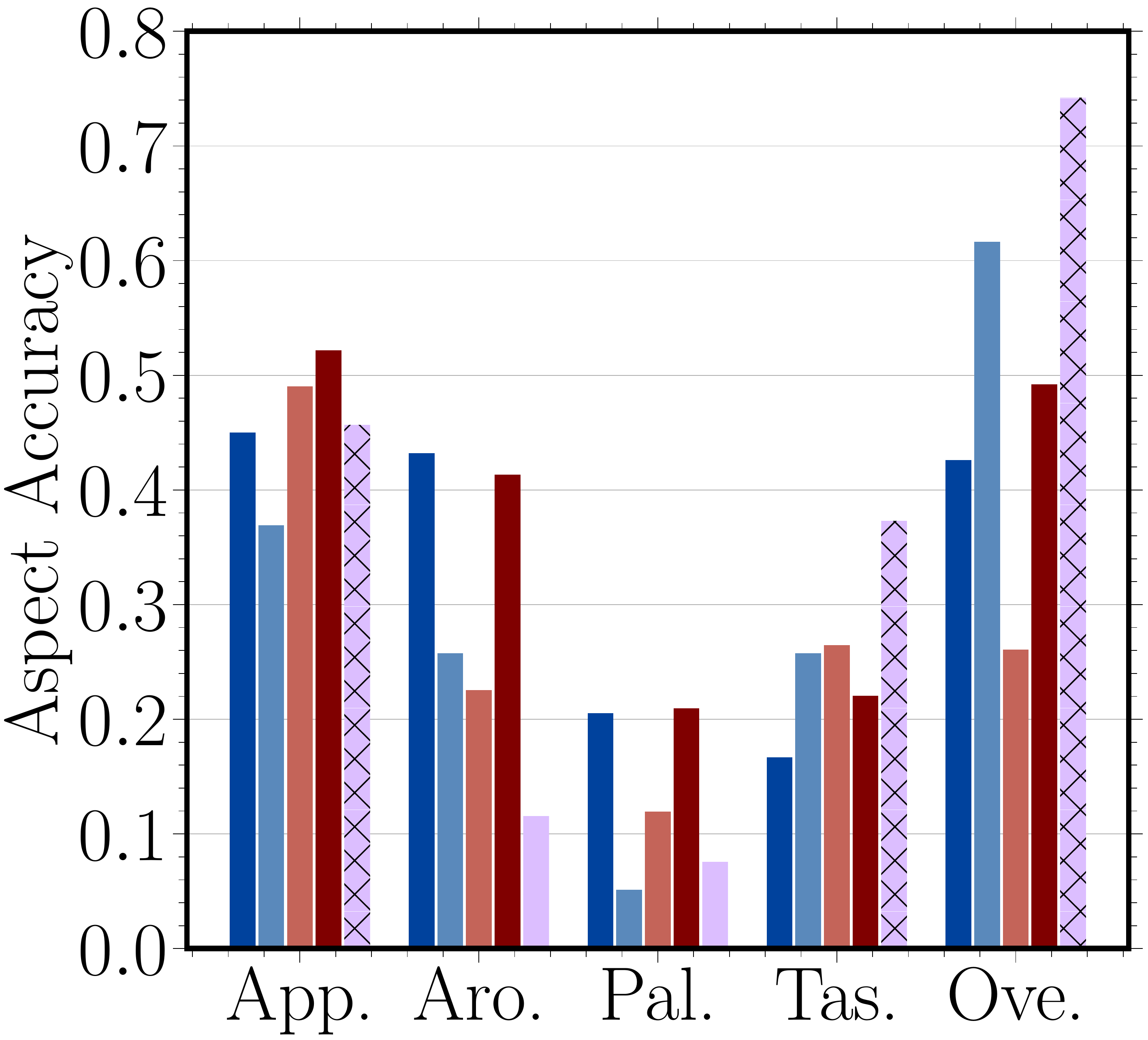}
\end{subfigure}
\par\bigskip
\par\bigskip

\begin{subfigure}[t]{0.325\textwidth}
\centering
\includegraphics[width=\textwidth]{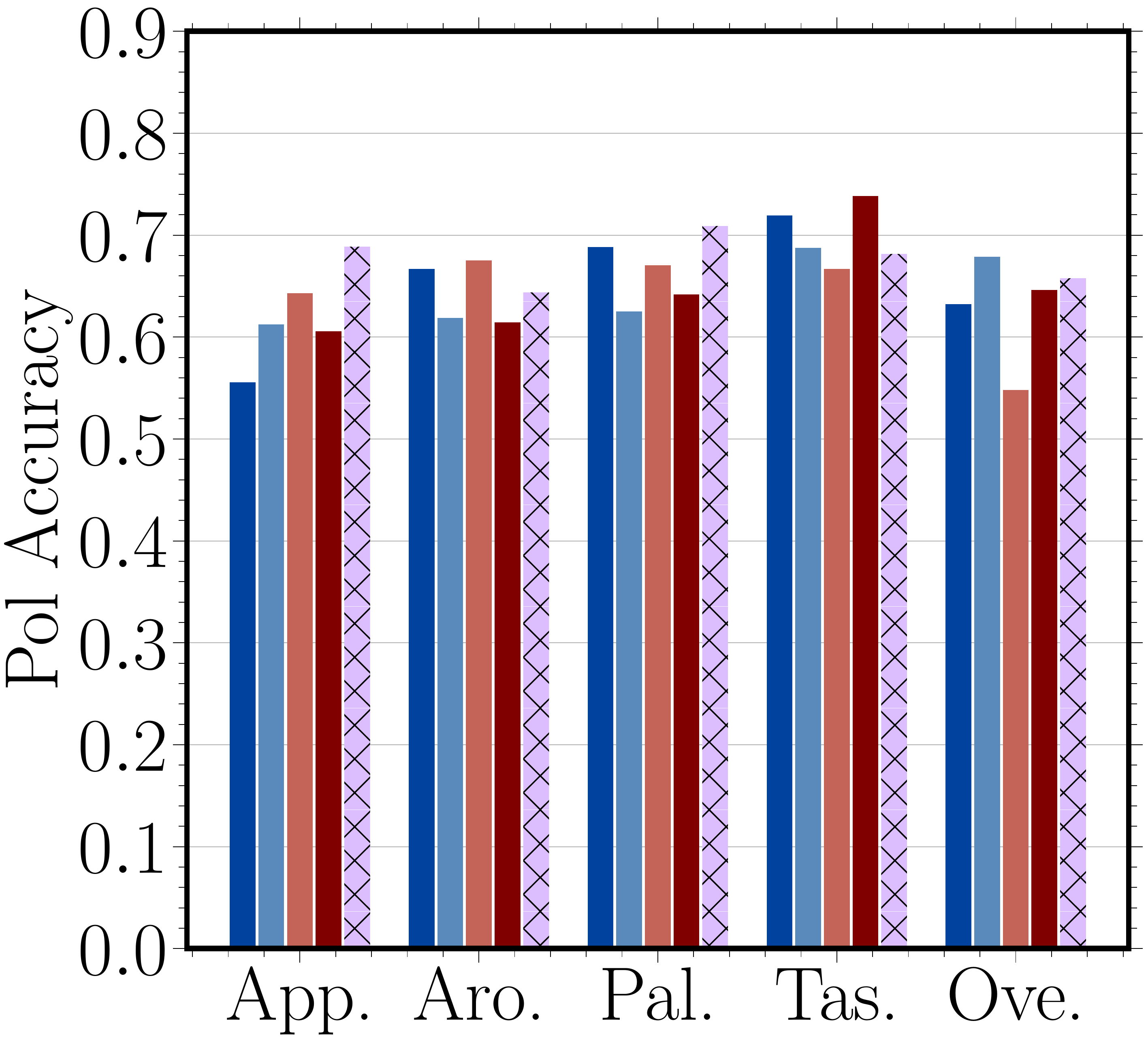}
\caption{Concept length $\ell=10$.}
\end{subfigure}
\begin{subfigure}[t]{0.325\textwidth}
\centering
\includegraphics[width=\textwidth]{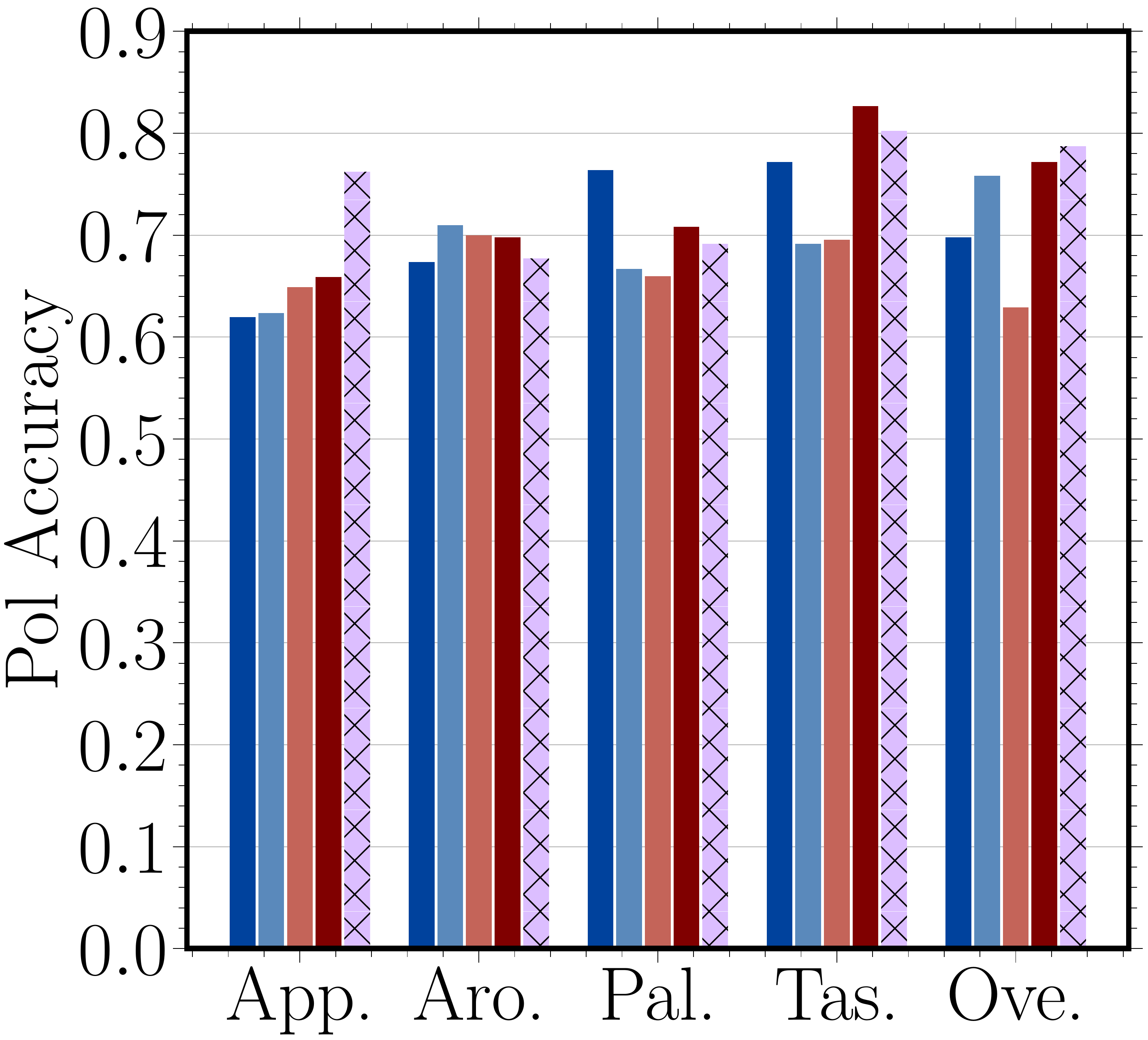}
\caption{Concept length $\ell=20$.}
\end{subfigure}
\begin{subfigure}[t]{0.325\textwidth}
\centering
\includegraphics[width=\textwidth]{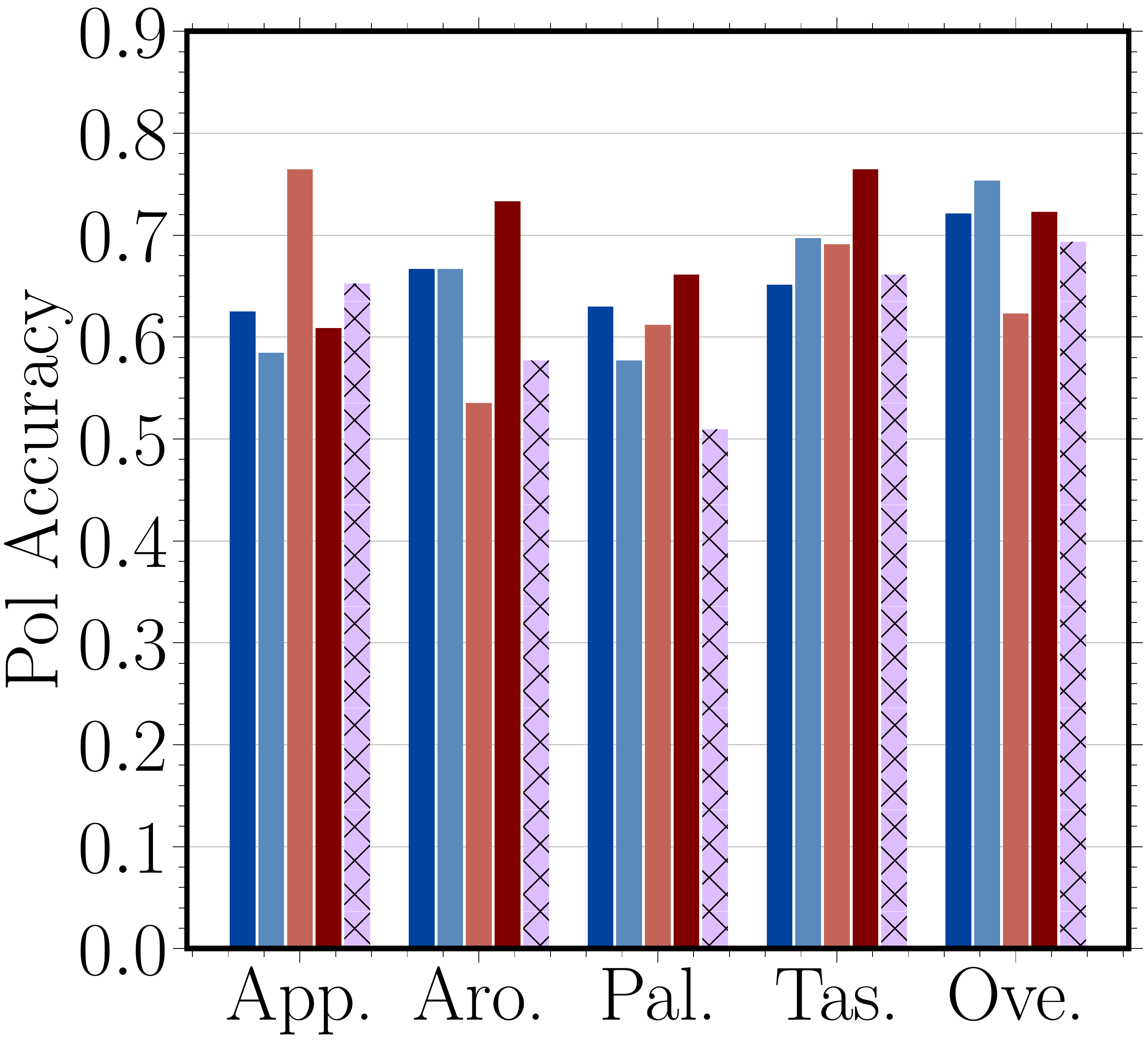}
\caption{Concept length $\ell=5$.}
\end{subfigure}

\caption{\label{fig_app_rq2_subj}Subjective performance per aspect of rationales for the multi-aspect beer reviews.}
\end{figure*}

All results (for the joint, the aspect, and the polarity accuracy) are shown in Figure~\ref{fig_app_rq2_subj}. In total, we used 7,500 samples ($100 \times 5 \times 5 \times 3$).

We also studied the error rates on each aspect. The Aroma and Palate aspects cause the highest error for all models. One possible reason is that users confuse these with the aspect Taste, hence their high correlations in rating scores \cite{antognini2019multi}.

\section{Extra Visualizations}
\label{app_samples}

\begin{figure*}[!t]
\centering
\begin{tabular}{@{}c@{}c@{}c@{}}
  \multicolumn{3}{c}{\includegraphics[width=0.45\textwidth,height=.4cm]{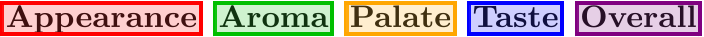}}\\
   ConRAT (Ours) & InvRAT \cite{chang2020invariant} & RNP \cite{lei-etal-2016-rationalizing}\\
     \includegraphics[width=0.36\textwidth,height=5.1cm]{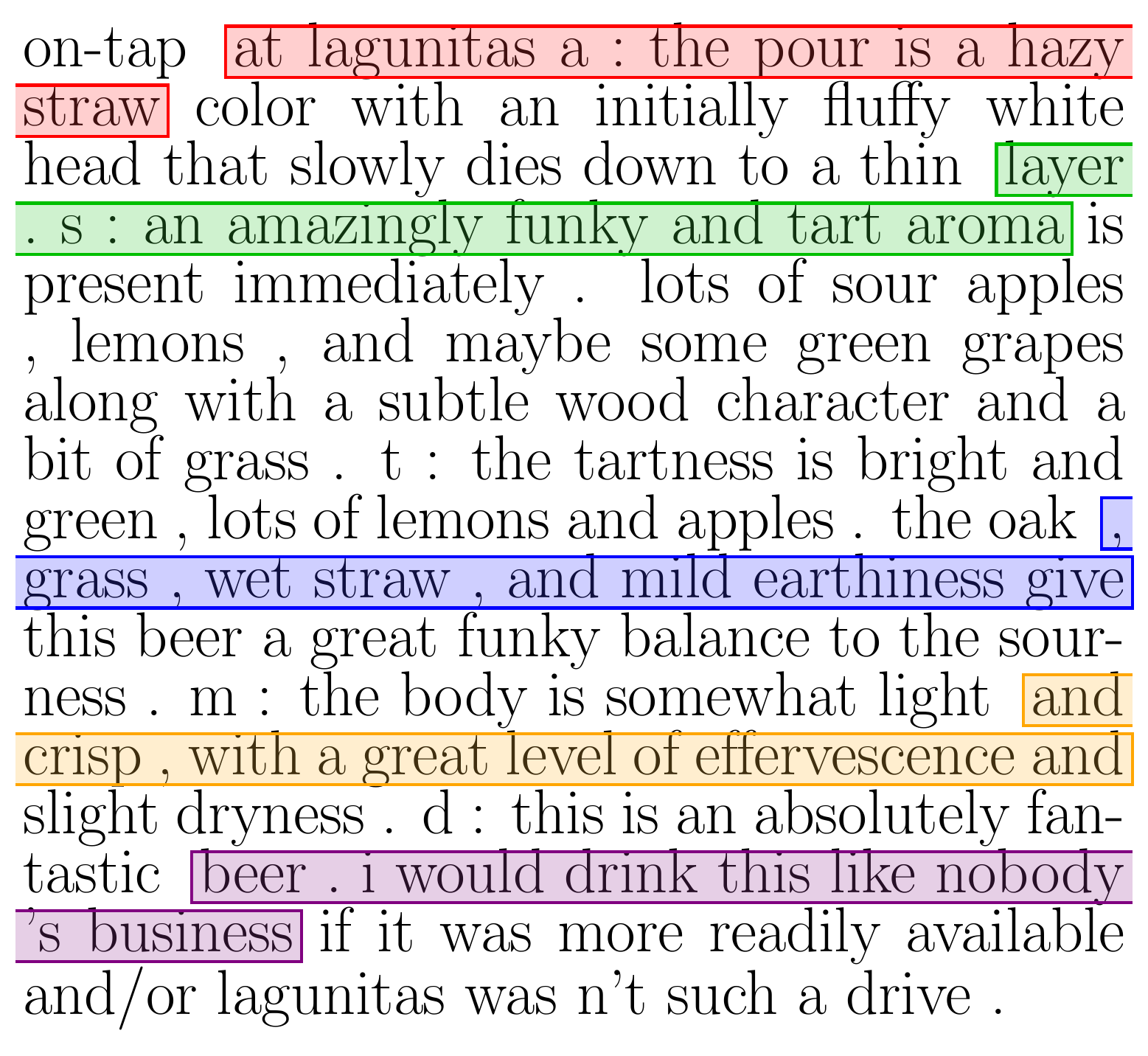}&     \includegraphics[width=0.36\textwidth,height=5.1cm]{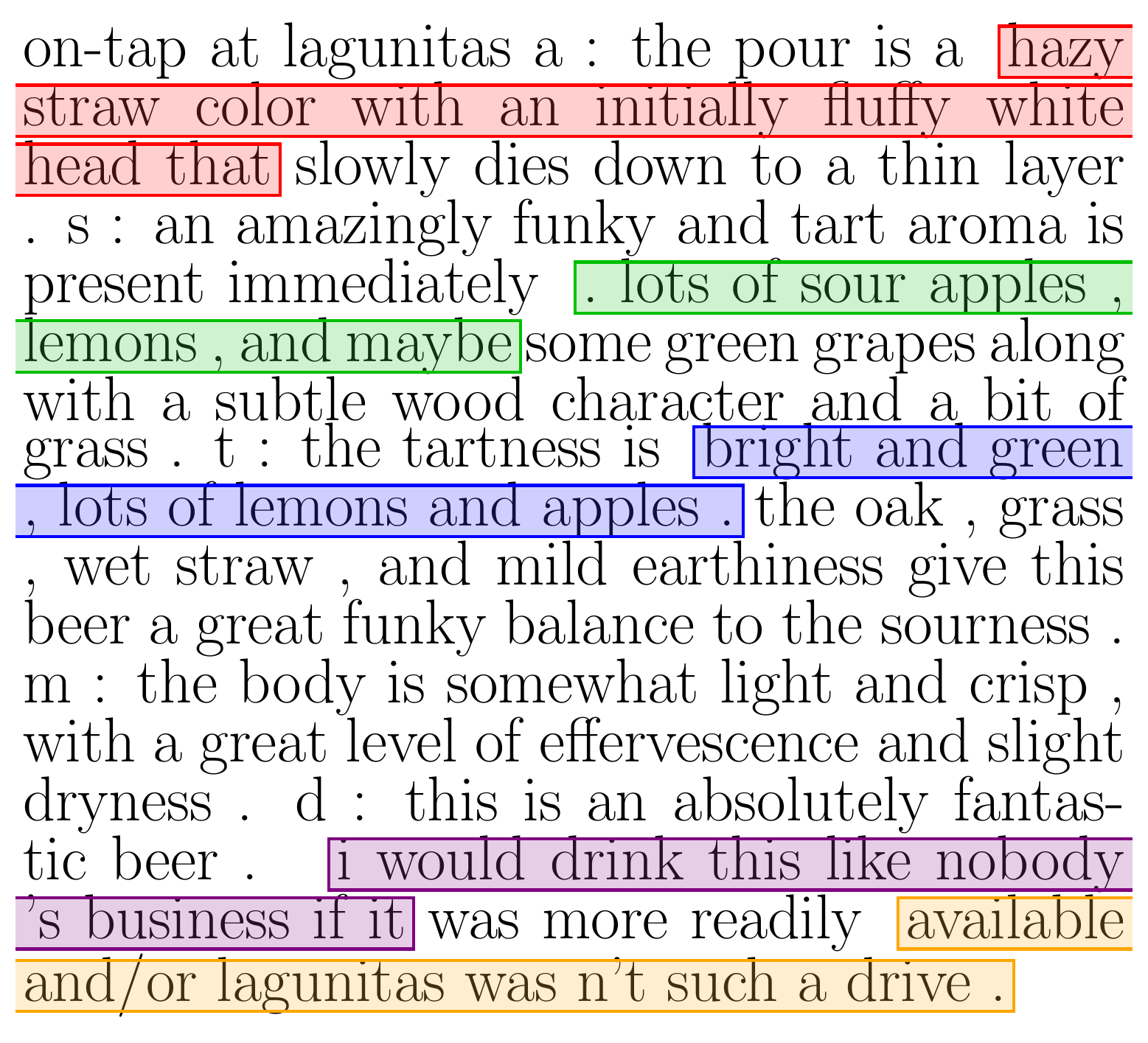}&     \includegraphics[width=0.36\textwidth,height=5.1cm]{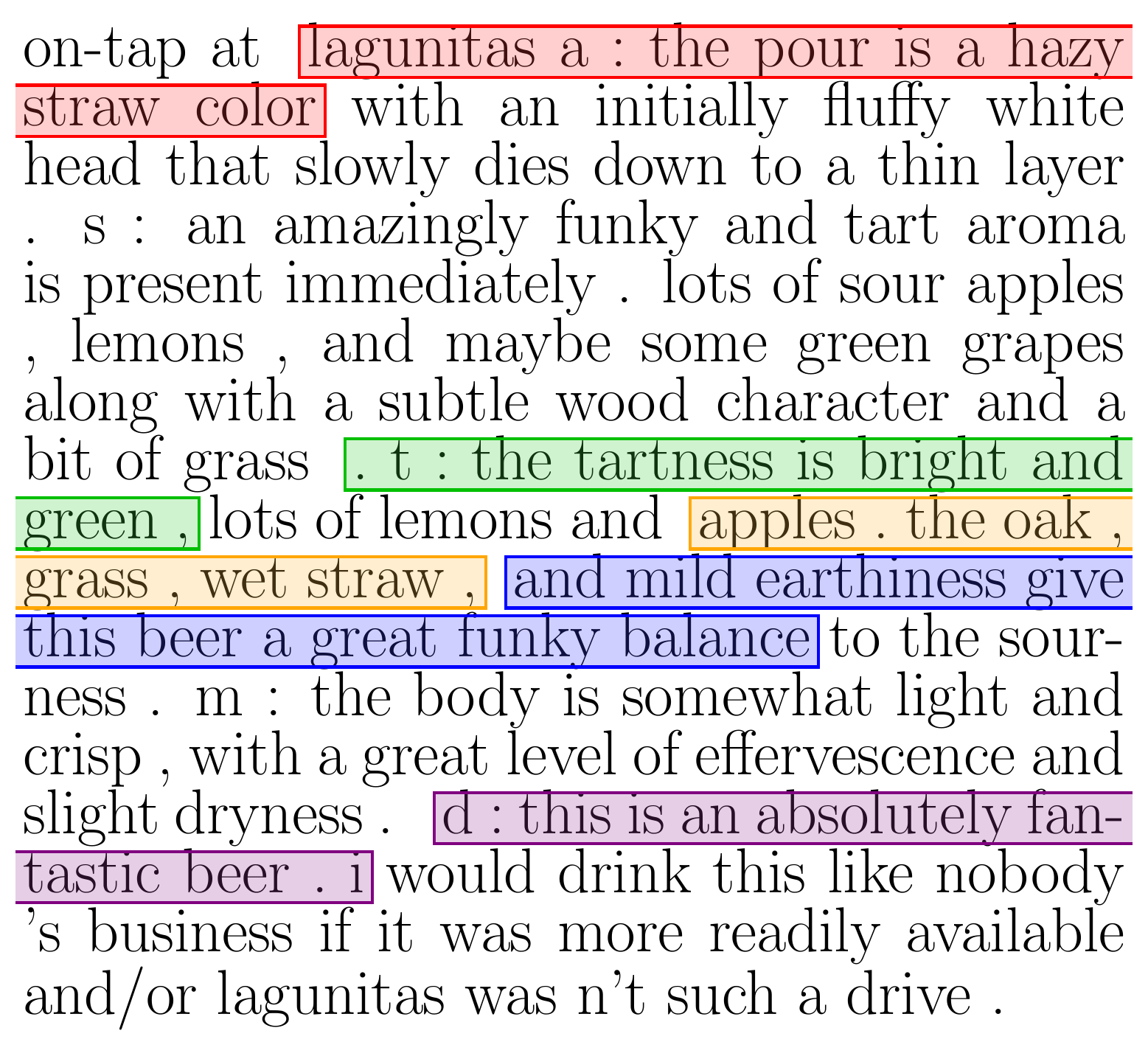}
\end{tabular}
\caption{\label{app_sample_10_1}Examples of generated rationales with $\ell=10$ for a beer review. \underline{Underline} highlights ambiguities.}

\begin{tabular}{@{}c@{}c@{}c@{}}
  \\
   ConRAT (Ours) & InvRAT \cite{chang2020invariant} & RNP \cite{lei-etal-2016-rationalizing}\\
     \includegraphics[width=0.36\textwidth,height=3.5cm]{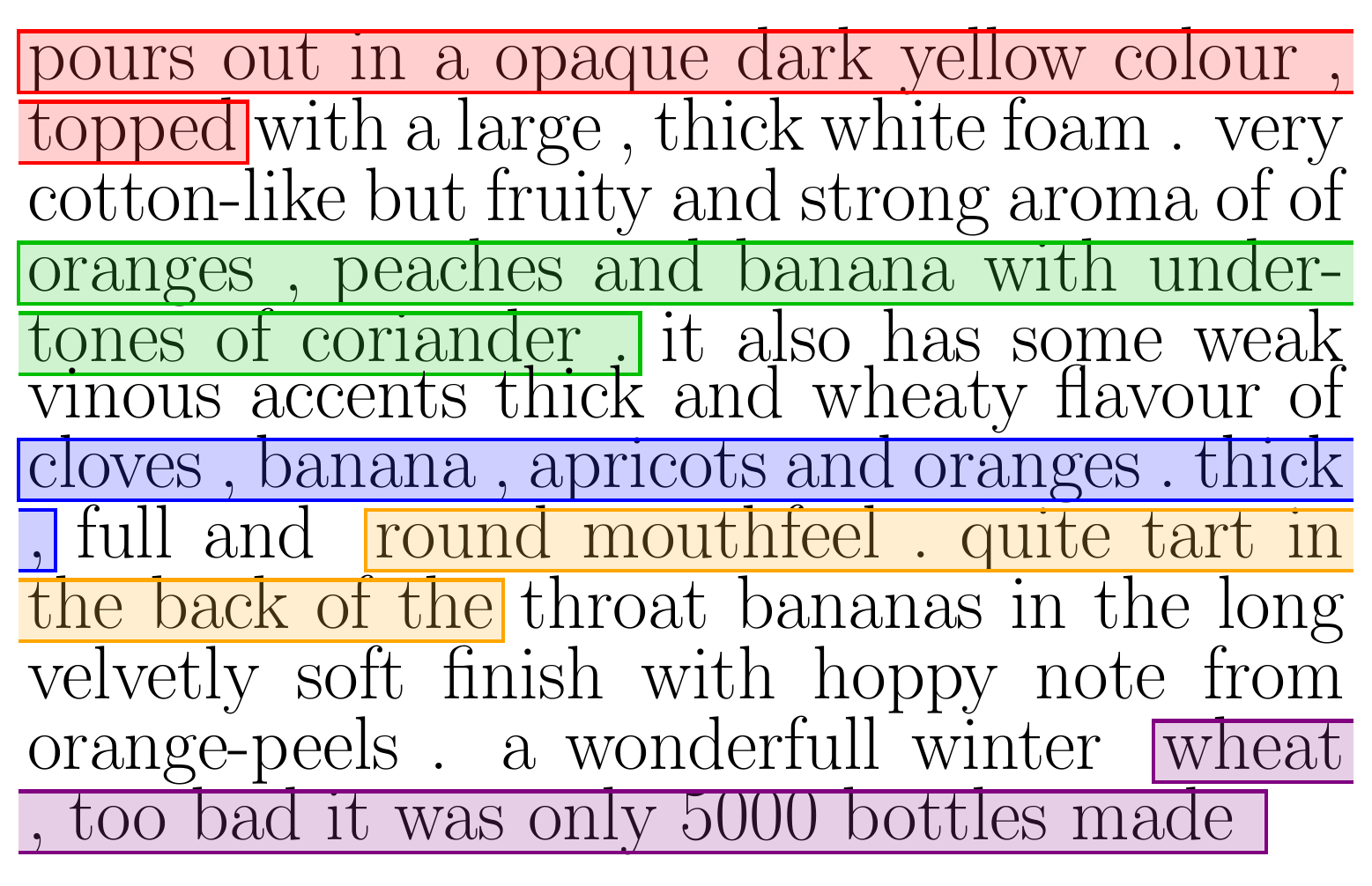}&     \includegraphics[width=0.36\textwidth,height=3.5cm]{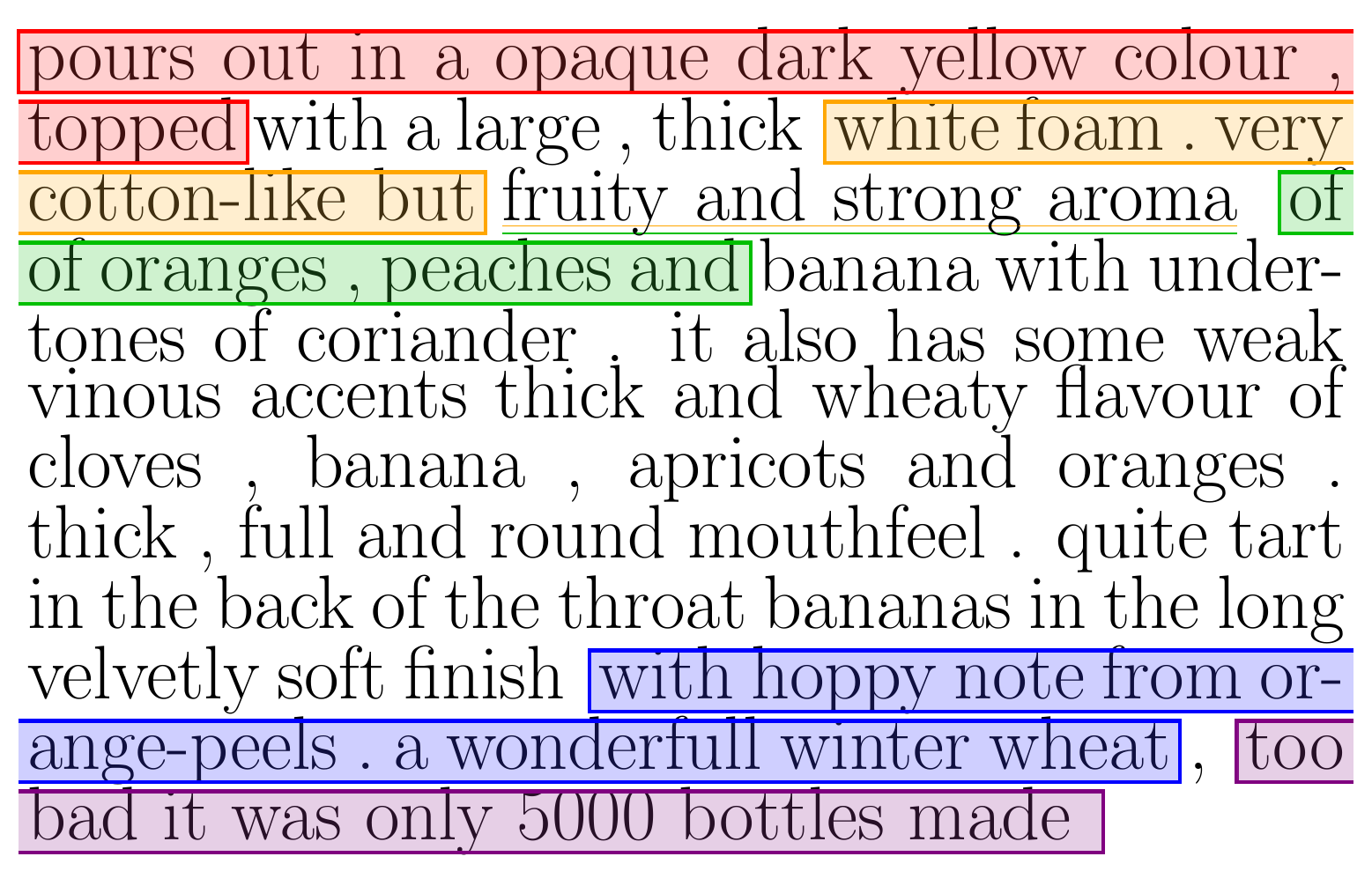}&     \includegraphics[width=0.36\textwidth,height=3.5cm]{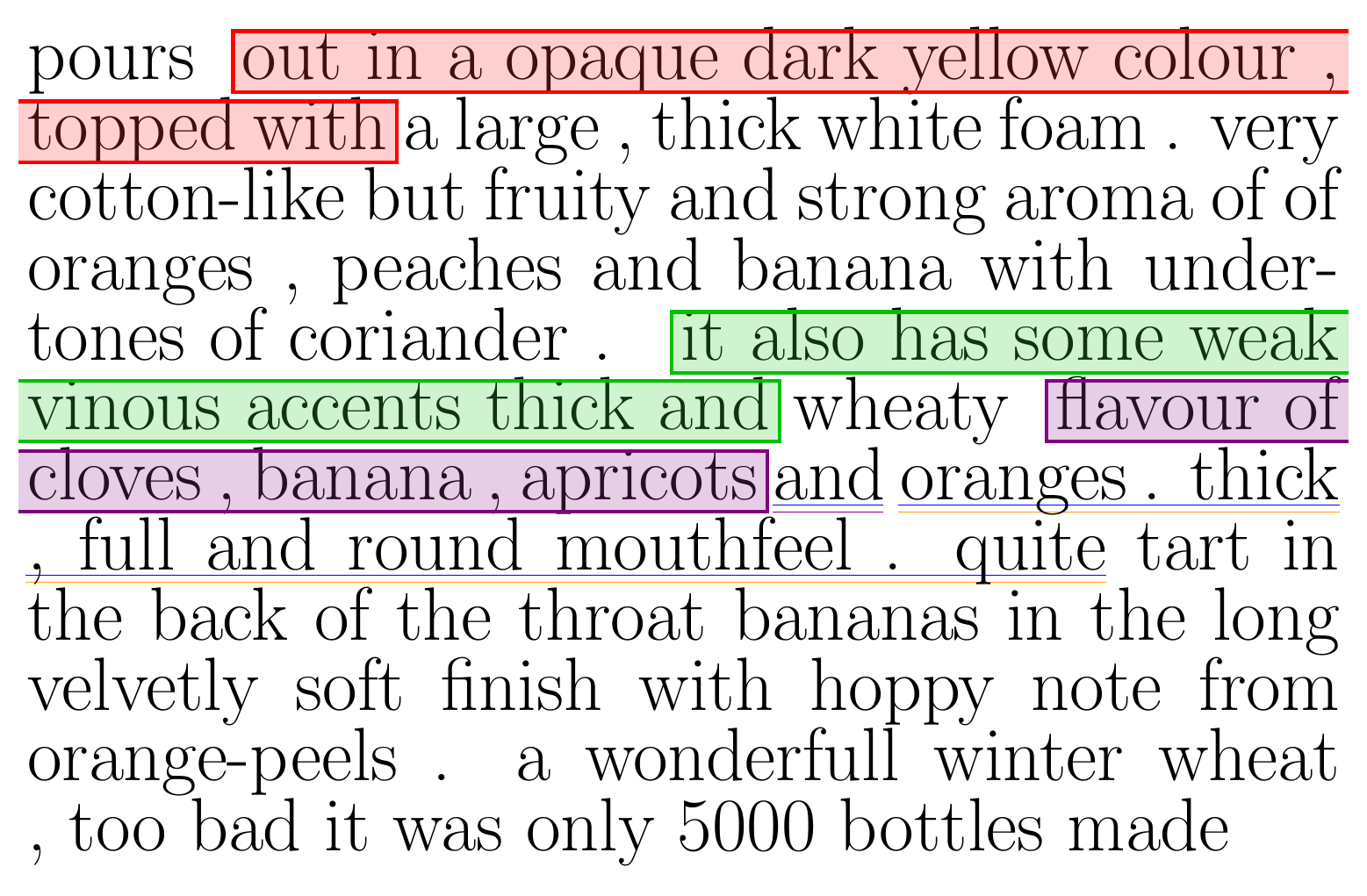}
\end{tabular}
\caption{\label{app_sample_10_2}Examples of generated rationales with $\ell=10$ for a beer review. \underline{Underline} highlights ambiguities.}

\begin{tabular}{@{}c@{}c@{}c@{}}
\\
   ConRAT (Ours) & InvRAT \cite{chang2020invariant} & RNP \cite{lei-etal-2016-rationalizing}\\
     \includegraphics[width=0.36\textwidth,height=3.5cm]{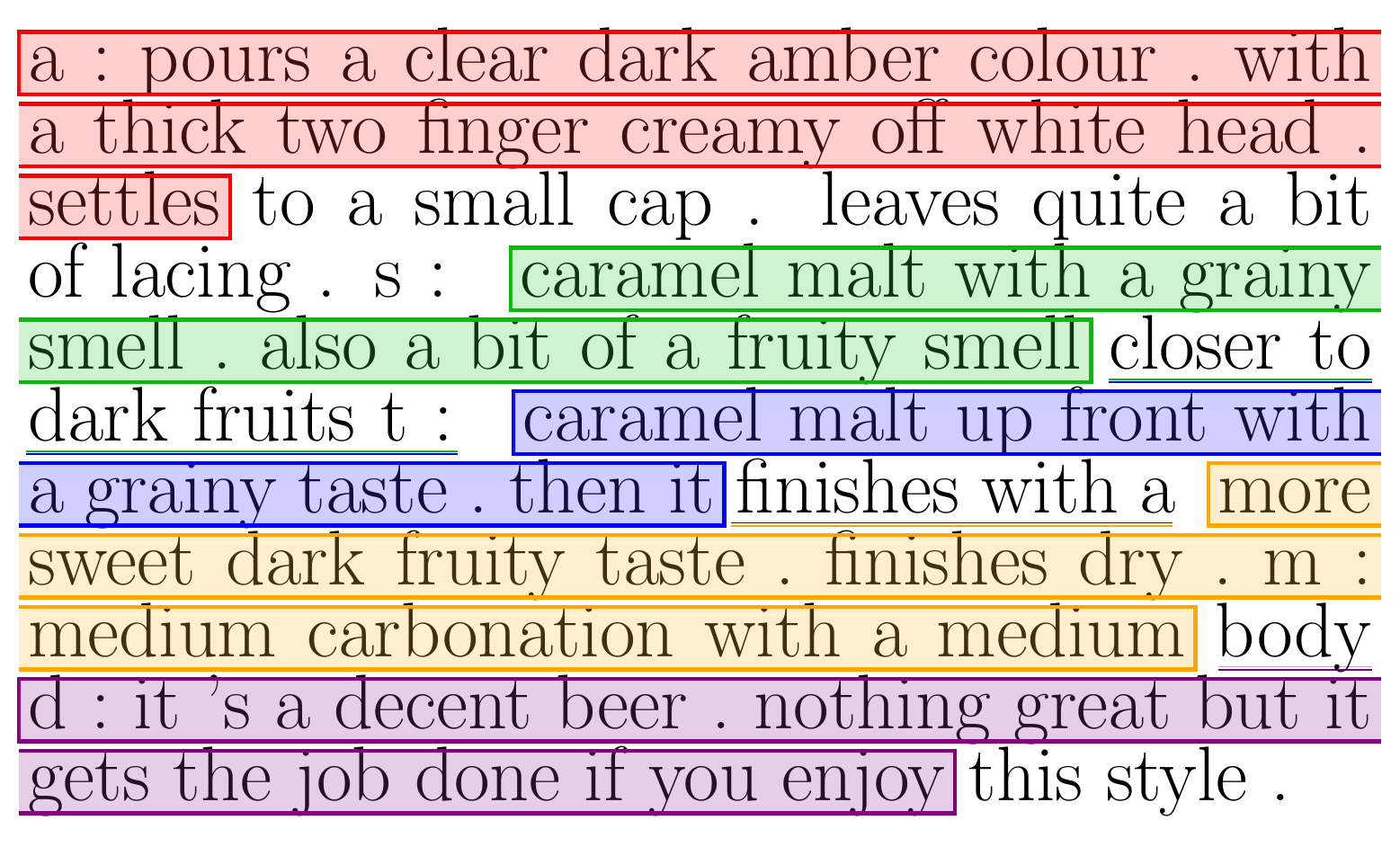}&     \includegraphics[width=0.36\textwidth,height=3.5cm]{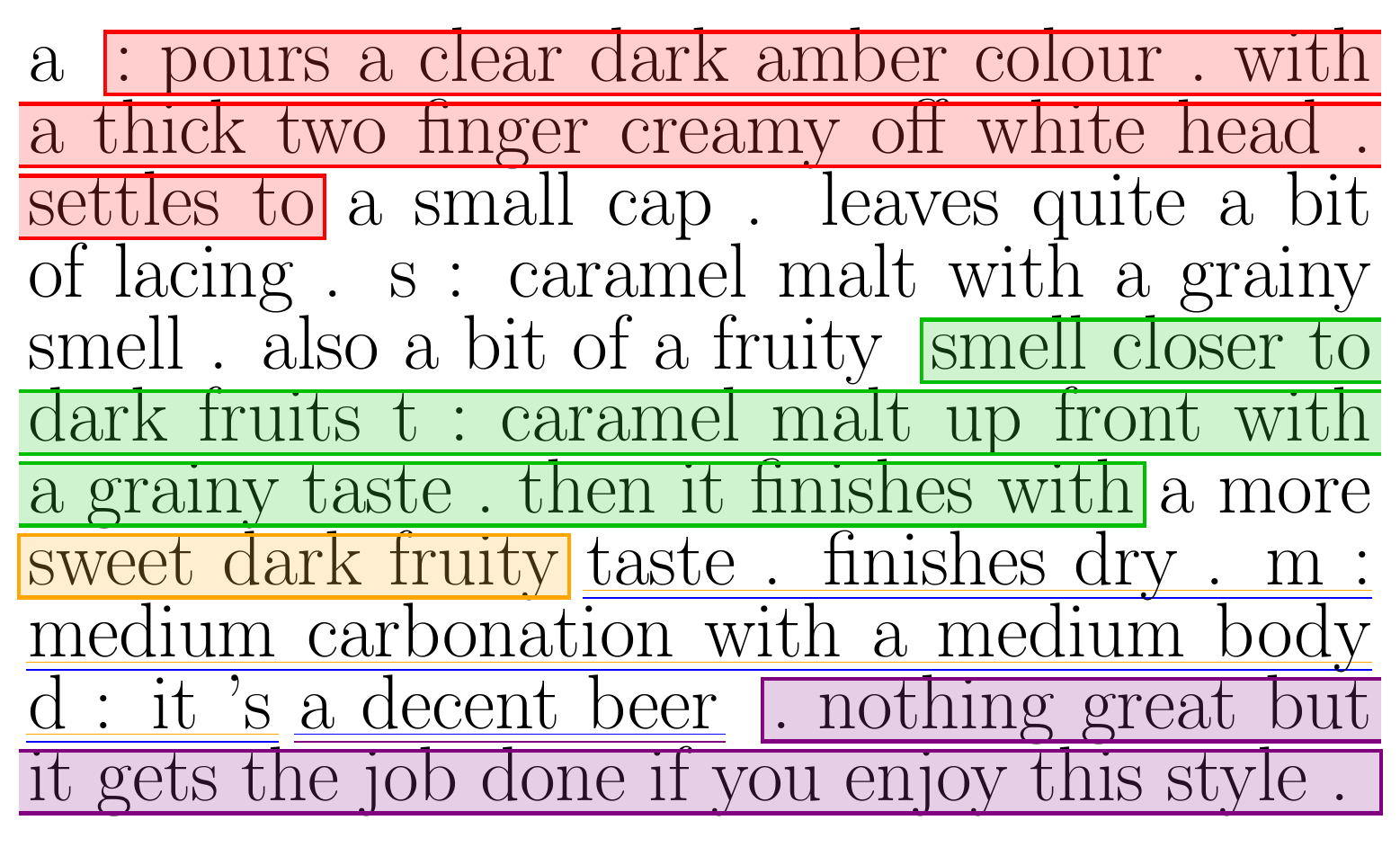}&     \includegraphics[width=0.36\textwidth,height=3.5cm]{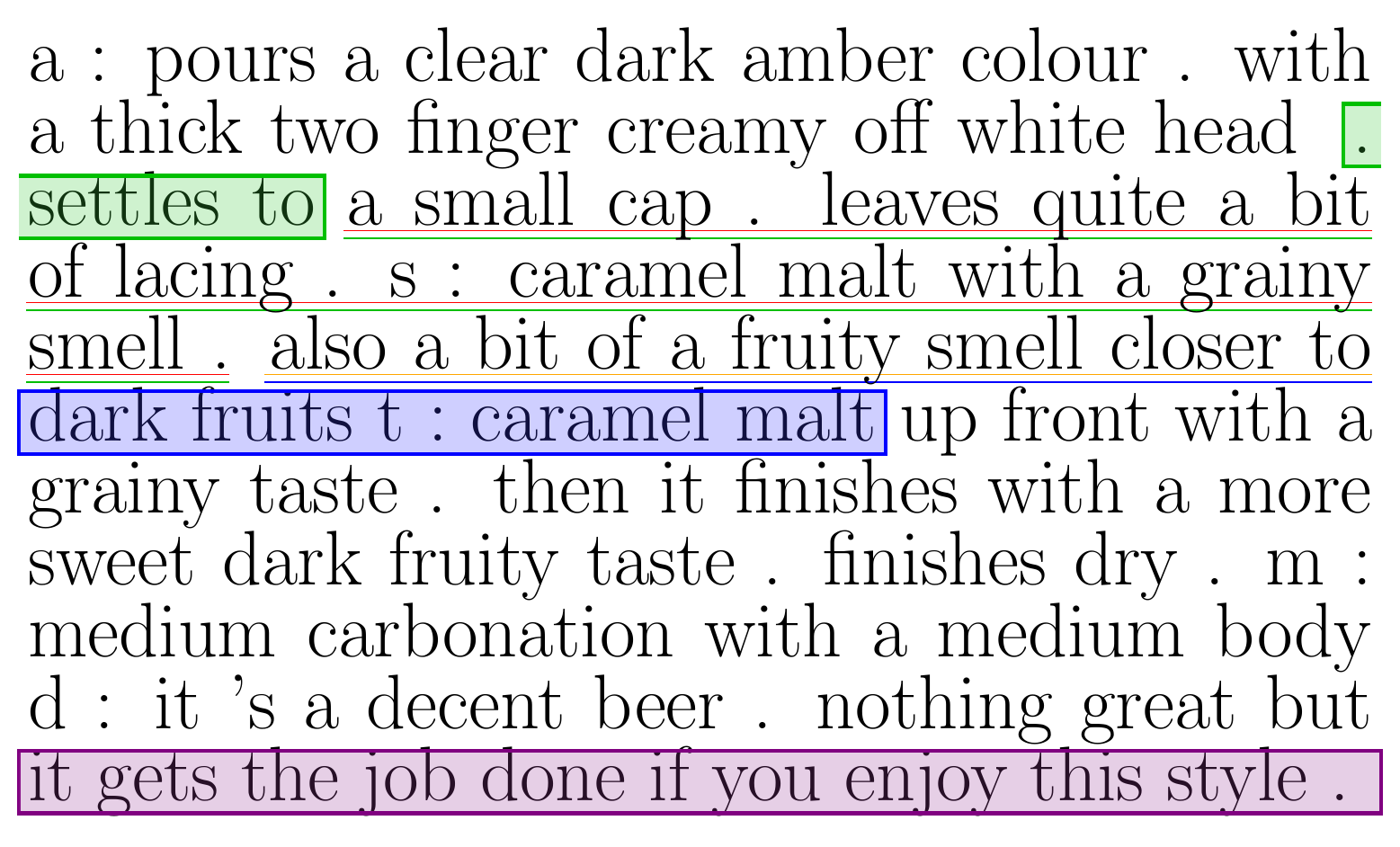}
\end{tabular}
\caption{\label{app_sample_20_1}Examples of generated rationales with $\ell=20$ for a beer review. \underline{Underline} highlights ambiguities.}

\begin{tabular}{@{}c@{}c@{}c@{}}
\\

   ConRAT (Ours) & InvRAT \cite{chang2020invariant} & RNP \cite{lei-etal-2016-rationalizing}\\
     \includegraphics[width=0.36\textwidth,height=4.5cm]{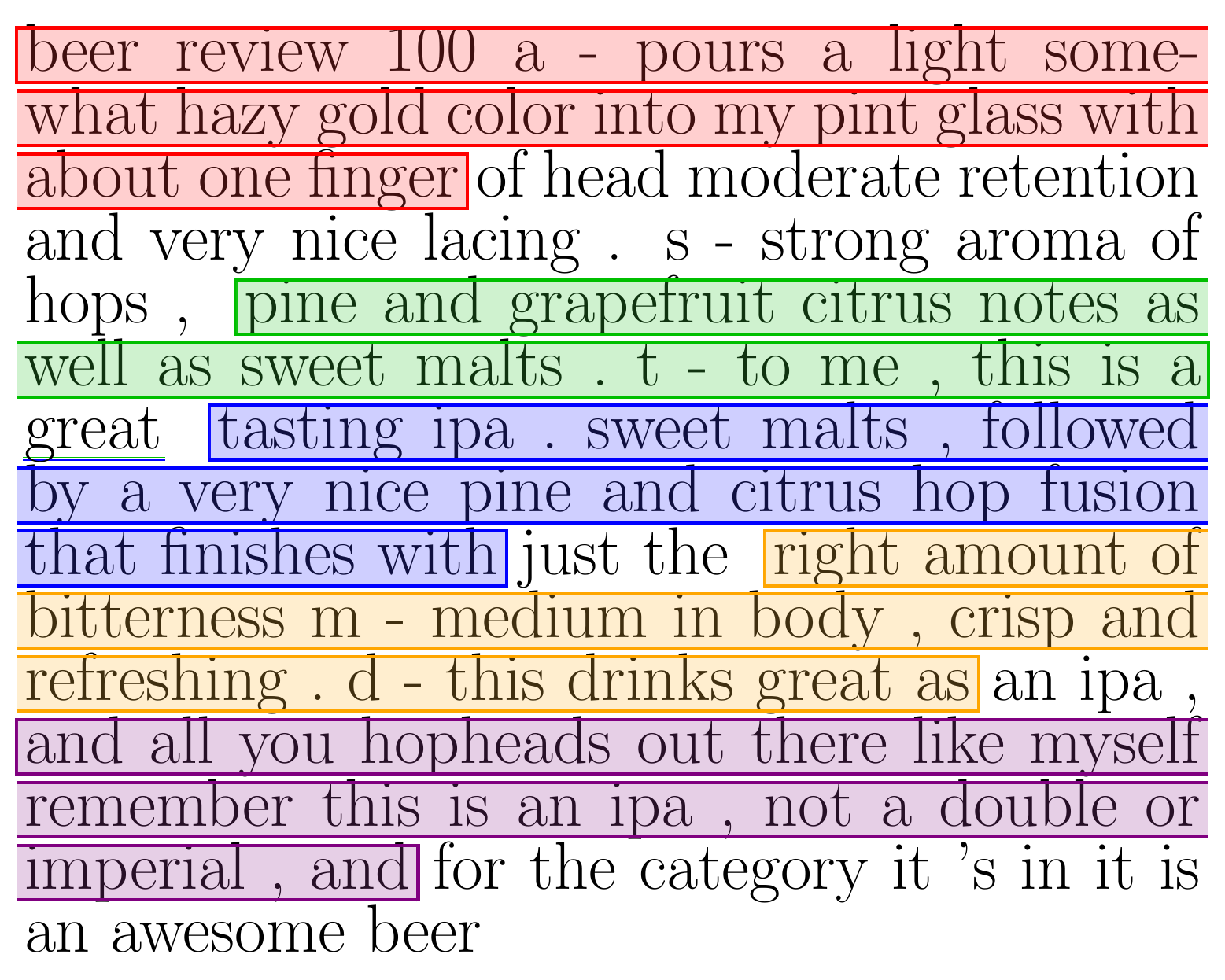}&     \includegraphics[width=0.36\textwidth,height=4.5cm]{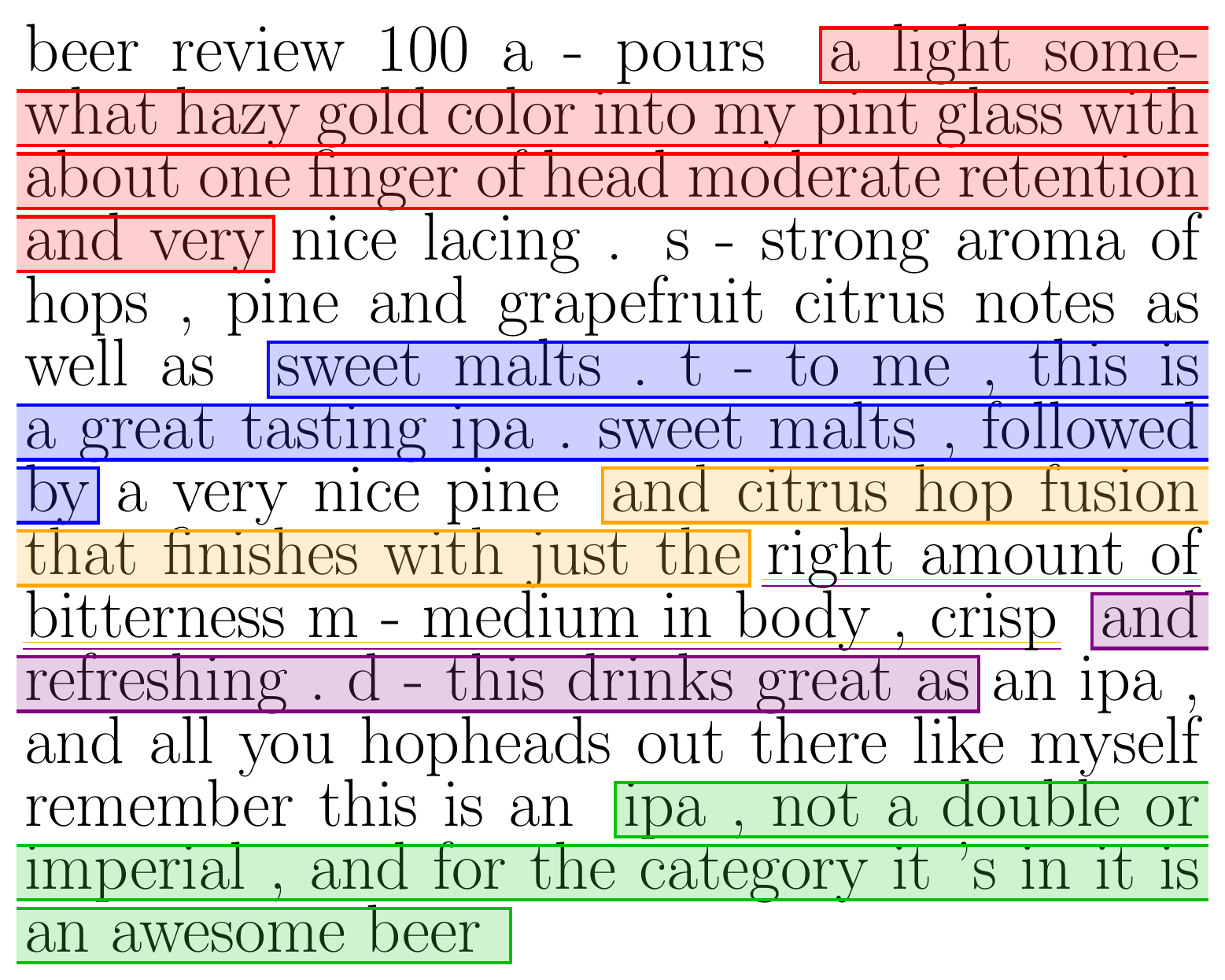}&     \includegraphics[width=0.36\textwidth,height=4.5cm]{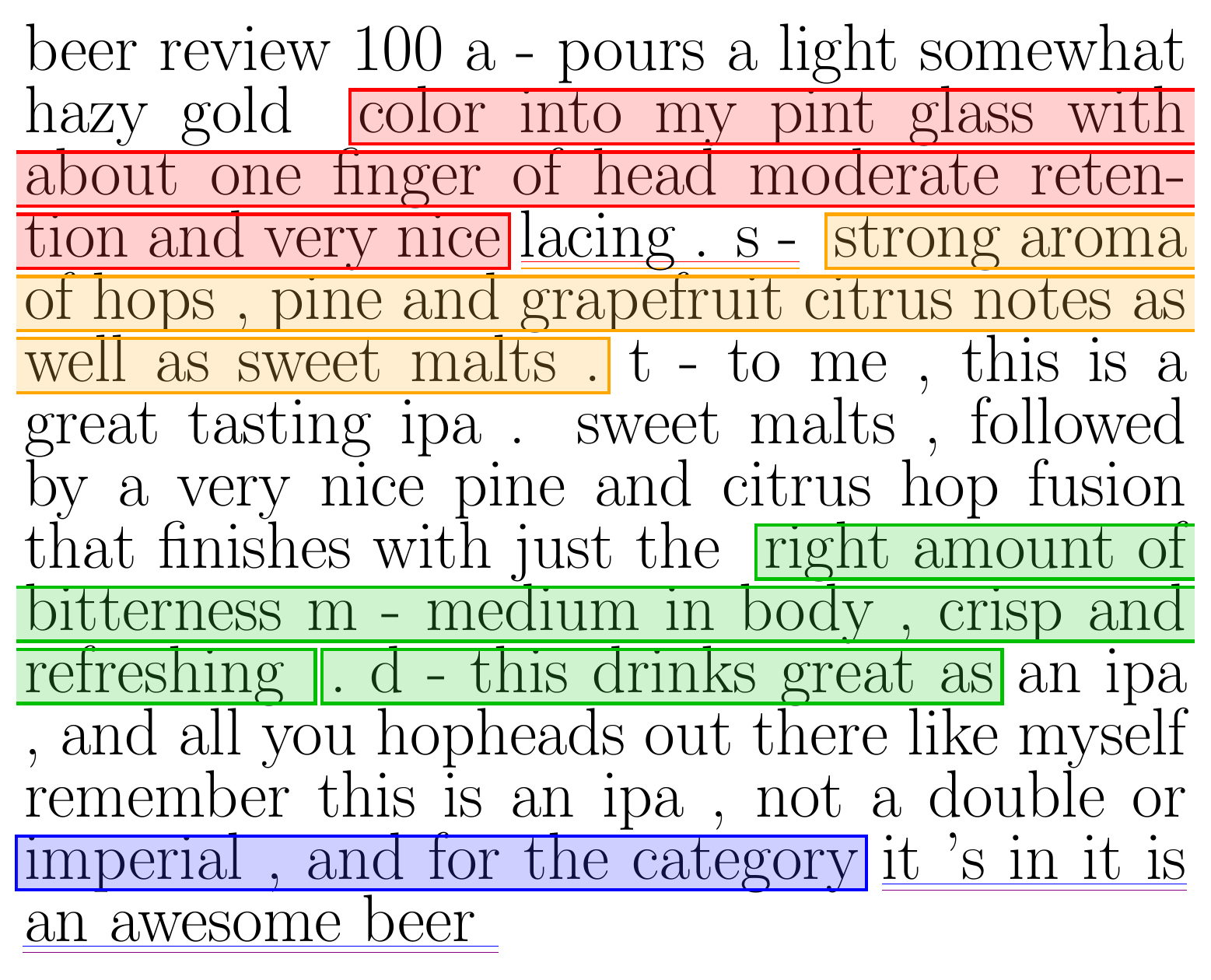}\\
\end{tabular}
\caption{\label{app_sample_20_2}Examples of generated rationales with $\ell=20$ for a beer review. \underline{Underline} highlights ambiguities.}
\end{figure*}

Additional samples of generated rationales are shown in Figure~\ref{app_sample_10_1}, \ref{app_sample_10_2}, \ref{app_sample_20_1}, and \ref{app_sample_20_2}. We can observe that baselines suffer from spurious correlations: the rationale for the aspect Aroma, Palate, and Taste are often exchanged, or several rationales pick the same text snippets. On the other hand, ConRAT finds better concepts while only trained on the overall aspect label. As it has been shown in prior work \cite{lei-etal-2016-rationalizing,chang2020invariant,antognini2019multi} rationale methods suffer from the high correlation between rating scores because each model is trained independently for each aspect. Therefore, they rely on the assumption that the data have low internal correlations, which does not reflect the real data distribution. By contrast, ConRAT alleviates this problem be finding all concepts in one training.

\end{document}